\documentclass{article}
\usepackage{ijcai18}

\usepackage{times}
\usepackage{xcolor}
\usepackage{soul}
\usepackage[utf8]{inputenc}
\usepackage[small]{caption}
\usepackage{times}
\usepackage{graphicx}
\usepackage{amsmath}
\usepackage{amssymb}
\usepackage{amsthm}
\usepackage{amsfonts}
\usepackage{mathrsfs}
\usepackage{latexsym}
\usepackage{bm}

\title{Salient Object Detection by Lossless Feature Reflection}

\iftrue
\author{
Pingping Zhang$^{1,3}$,
Wei Liu$^{2,3}$,
Huchuan Lu$^1$,
Chunhua Shen$^3$
\\
$^1$ Dalian University of Technology, Dalian, 116024, P.R. China \\
$^2$ Shanghai Jiao Tong University, Shanghai, 200240, P.R. China \\
$^3$ University of Adelaide, Adelaide, SA 5005, Australia\\
jssxzhpp@mail.dlut.edu.cn;
liuwei.1989@sjtu.edu.cn;
lhchuan@dlut.edu.cn;
chunhua.shen@adelaide.edu.au
}
\fi

\begin{document}

\maketitle

\begin{abstract}
Salient object detection, which aims to identify and locate the most salient pixels or regions in images, has been attracting more and more interest due to its various real-world applications.
However, this vision task is quite challenging, especially under complex image scenes.
Inspired by the intrinsic reflection of natural images, in this paper we propose a novel feature learning framework for large-scale salient object detection.
Specifically, we design a symmetrical fully convolutional network (SFCN) to learn complementary saliency features under the guidance of lossless feature reflection.
The location information, together with contextual and semantic information, of salient objects are jointly utilized to supervise the proposed network for more accurate saliency predictions.
In addition, to overcome the blurry boundary problem, we propose a new structural loss function to learn clear object boundaries and spatially consistent saliency.
The coarse prediction results are effectively refined by these structural information for performance improvements.
Extensive experiments on seven saliency detection datasets demonstrate that our approach achieves consistently superior performance and outperforms the very recent state-of-the-art methods.
\end{abstract}

\section{Introduction}
As a fundamental yet challenging task in computer vision, salient object detection (SOD) aims to identify and locate distinctive objects or regions which attract human attention in natural images.
In general, SOD is regarded as a prerequisite step to narrow down subsequent object-related vision tasks.
%
For example, it can be used in image retrieval, sematic segmentation, visual tracking and person re-identification, etc.

In the past two decades, a large number of SOD methods have been proposed.
%
Most of them have been well summarized in~\cite{borji2015salient}.
According to that work, conventional SOD methods focus on extracting discriminative local and global handcrafted features from pixels or regions to represent their visual properties.
With several heuristic priors, these methods predict salient scores according to the extracted features for saliency detection.
Although great success has been made, there still exist many important problems which need to be solved.
For example, the low-level handcrafted features suffer from limited representation capability, and are difficult to capture the semantic and structural information of objects in images, which is very important for more accurate SOD.
What's more, to further extract powerful and robust visual features manually is a tough mission for performance improvement, especially in complex image scenes, such as cluttered backgrounds and low-contrast imaging patterns.

With the recent prevalence of deep architectures, many remarkable progresses have been achieved in a wide range of computer vision tasks, \emph{e.g.}, image classification~\cite{simonyan2014very}, object detection~\cite{girshick2014rich}, and semantic segmentation~\cite{long2015fully}.
Thus, many researchers start to make their efforts to utilize deep convolutional neural networks (CNNs) for SOD and have achieved favourable performance, since CNNs have strong ability to automatically extract high-level feature representations, successfully avoiding the drawbacks of handcrafted features.
However, most of state-of-the-art SOD methods still require large-scale pre-trained CNNs, which usually employ the strided convolution and pooling operations.
These downsampling methods increase the receptive field of CNNs, helping to extract high-level semantic features, nevertheless they inevitably drop the location information and fine details of objects, leading to unclear boundary predictions.
Furthermore, the lack of structural supervision also makes SOD an extremely challenging problem in complex image scenes.

\begin{figure}
\begin{center}
\hspace{-4mm}
\includegraphics[width=1.04\linewidth,height=5cm]{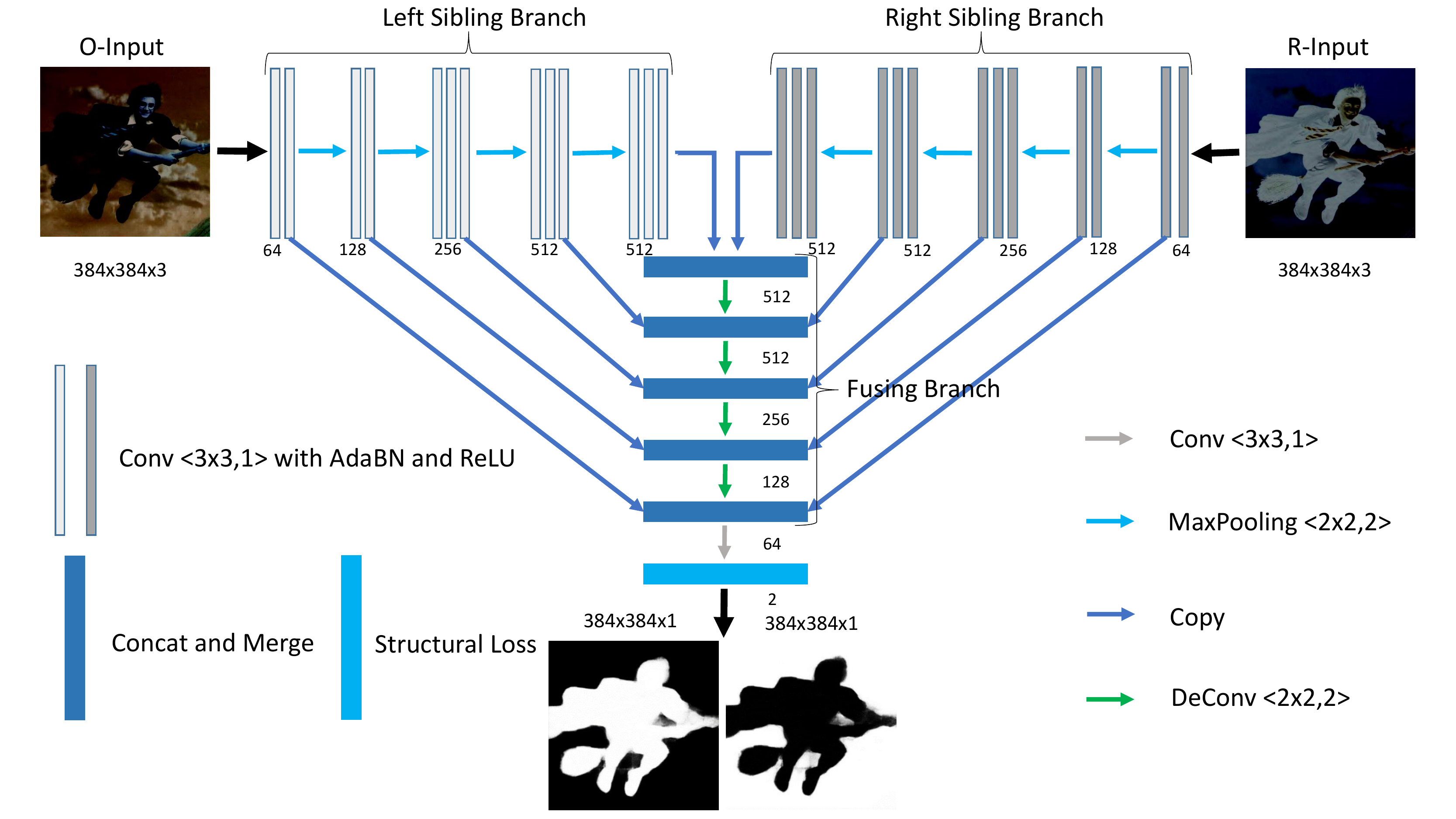}
\end{center}
\vspace{-4mm}
\caption{The semantic overview of our proposed SFCN.}
\label{fig:framework}
\vspace{-4mm}
\end{figure}

In order to utilize the semantic and structural information derived from deep pre-trained CNNs, we propose to solve both tasks of complementary feature extraction and saliency region classification with an unified framework which is learned in the end-to-end manner.
Specifically, we design a symmetrical fully convolutional network (SFCN) architecture which consists of two sibling branches and one fusing branch, as illustrated in Fig.~\ref{fig:framework}.
The two sibling branches take reciprocal image pairs as inputs and share weights for learning complementary visual features under the guidance of lossless feature reflection.
The fusing branch integrates the multi-level complementary features in a hierarchical manner for SOD.
More importantly, to effectively train our network, we propose a novel loss function which incorporates structural information and supervises the three branches during the training process.
In this manner, our proposed model sufficiently captures the boundaries and spatial contexts of salient objects, hence significantly boosts the performance of SOD.

In summary, \textbf{our contributions} are three folds:
\begin{itemize}
\vspace{-0.5mm}
\item
We present a novel network architecture, \emph{i.e.}, SFCN, which is symmetrically designed to learn complementary visual features and predict accurate saliency maps under the guidance of lossless feature reflection.
\vspace{-0.5mm}
\item
We propose a new structural loss function to learn clear object boundaries and spatially consistent saliency.
This loss function is able to utilize the location, contextual and semantic information of salient objects to supervise the proposed SFCN for performance improvements.
\vspace{-0.5mm}
\item
Extensive experiments on seven large-scale saliency benchmarks demonstrate that the proposed approach achieves superior performance and outperforms the very recent state-of-the-art methods by a large margin.
\end{itemize}
\section{Related Work}
%
%
{\flushleft\textbf{Salient Object Detection.}} Recent years, deep learning based methods have achieved solid performance improvements in SOD.
For example,~\cite{wang2015deep} integrate both local pixel estimation and global proposal search for SOD by training two deep neural networks.
~\cite{zhao2015saliency} propose a multi-context deep CNN framework to benefit from the local context and global context of salient objects.
~\cite{li2015visual} employ multiple deep CNNs to extract multi-scale features for saliency prediction.
Then they propose a deep contrast network to combine a pixel-level stream and segment-wise stream for saliency estimation~\cite{li2016deep}.
Inspired by the great success of fully convolutional networks (FCNs)~\cite{long2015fully},~\cite{wang2016saliency} develop a recurrent FCN to incorporate saliency priors for more accurate saliency map inference.
~\cite{liu2016dhsnet} also design a deep hierarchical network to learn a coarse global estimation and then refine the saliency map hierarchically and progressively.
Then,~\cite{hou2017deeply} introduce dense short connections to the skip-layers within the holistically-nested edge detection (HED) architecture~\cite{xie2015holistically} to get rich multi-scale features for SOD.
~\cite{zhang2017amulet} propose a bidirectional learning framework to aggregate multi-level convolutional features for SOD.
And they also develop a novel dropout to learn the deep uncertain convolutional features to enhance the robustness and accuracy of saliency detection~\cite{zhang2017learning}.
~\cite{wang2017stagewise} provide a stage-wise refinement framework to gradually get accurate saliency detection results.
Despite these approaches employ powerful CNNs and make remarkable success in SOD, there still exist some obvious problems.
For example, the strategies of multiple-stage training reduce the efficiency.
And the explicit pixel-wise loss functions used by these methods for model training cannot well reflect the structural information of salient objects.
Hence, there is still a large space for performance improvements.
{\flushleft\textbf{Image Intrinsic Reflection.}} Image intrinsic reflection is a classical topic in computer vision field.
It aims to separate a color image into two intrinsic reflection images: an image of just the highlights, and the original image with the highlights removed.
It can be used to segment and analyze surfaces with image color variations. 
Most of existing methods are based on the Retinex model~\cite{land1971lightness}, which captures image information for Mondrian images: images of a planar canvas that is covered by samll patches of constant reflectance and illuminated by multiple light sources.
Recent years, researchers have augmented the basic Retinex model with non-local texture cues~\cite{zhao2012closed} and sparsity priors~\cite{shen2011intrinsic}.
Sophisticated techniques that recover reflectance and shading along with a shape estimate have also been proposed~\cite{barron2012color}.
Inspired by these works, we construct a reciprocal image pair based on the input image (see Section 3.1).
However, there are three obvious differences between our method and previous intrinsic reflection methods:
1) the objective is different. The aim of previous methods is explaining an input RGB image by estimating albedo and shading fields. Our aim is to learn complementary visual features for SOD.
2) the resulting image pair is different. Image intrinsic reflection methods usually factory an input image into a reflectance image and a shading image, while our method builds a reciprocal image pair for each image as the input of deep networks.
3) the source of reflection is different. The source of previous intrinsic reflection methods is the albedo of depicted
surfaces, while our reflection is originated from deep features in CNNs. Therefore, our reflection is feature-level not image-level.
%
\section{The Proposed Method}
Fig.~\ref{fig:framework} illustrates the semantic overview of our method.
We first convert an input RGB image into a reciprocal image pair, including the origin image (O-Input) and the reflection image (R-Input), by utilizing the ImageNet mean~\cite{deng2009imagenet} and a pixel-wise negation operator.
Then the image pair is fed into the sibling branches of our proposed SFCN, extracting multi-level deep features.
Afterwards, the fusing branch hierarchically integrates the complementary features into the same resolution of input images.
Finally, the saliency map is predicted by exploiting integrated features and the structural loss.
In the following subsections, we elaborate the proposed SFCN architecture and the weighted structural loss.
\subsection{Symmetrical FCN}
The proposed SFCN is an end-to-end fully convolutional network.
It consists of three main branches with a paired reciprocal image input to achieve lossless feature reflection learning.
%
%
{\flushleft\textbf{Reciprocal Image Input.}}
To capture complementary image information, we first convert the given RGB image $X\in R^{W\times H\times 3}$ to a reciprocal image pair by the following reflection function,
\begin{align}
Rec(X,k) &= (X-M,k(M-X))),\\
&= (X-M,-k(X-M))\\
&= (X_O, X^{k}_R).
  \label{equ:equ1}
\end{align}
where $k$ is a hyperparameter to control the reflection scale and $M\in R^{W\times H\times 3}$ is the mean of an image or image dataset.
From above equations, one can see that the converted image pair, \emph{i.e.}, $X_O$ and $X^{k}_R$, is reciprocal with a reflection plane.
In detail, the reflection scheme is a pixel-wise negation operator, allowing the given images to be reflected in both positive and negative directions while maintaining the same content of images.
%
%
In the proposed reflection, we use the multiplicative operator to measure the reflection scale, but it is not the only feasible method.
For example, this reflection can be combined with other non-linear operators, such as quadratic form, to add more diversity.
To reduce the computation burden, we use $k=1$ and the well-known mean of the ImageNet dataset.
{\flushleft\textbf{Sibling Branches with AdaBN.}}
Based on the reciprocal image pair, we propose two sibling branches to extract complementary reflection features.
More specifically, we build each sibling branch, following the VGG-16 model~\cite{simonyan2014very}.
Each sibling branch has 13 convolutional layers (kernel size = $3\times 3$, stride size = 1) and 4 max pooling layers (pooling size = $2\times 2$, stride = 2).
To achieve the lossless reflection features, the two sibling branches are designed to share weights in convolutional layers, but with adaptive batch normalization (AdaBN).
In other words, we keep the weights of corresponding convolutional layers of the two sibling branches the same, while use different learnable BN between the convolution and ReLU operators~\cite{zhang2017amulet}.
The main reason of this design is that after the reflection transform, the reciprocal images have different image domains. %
Domain related knowledge heavily affects the statistics of BN layers.
In order to learn domain invariant features, it's beneficial for each domain to keep its own BN statistics in each layers.
{\flushleft\textbf{Hierarchical Feature Fusion.}}
After extracting multi-level reflection features, we adhere an additional fusing branch to integrate them for the saliency prediction.
In order to preserve the spatial structure and enhance the contextual information, we integrate the multi-level reflection features in a hierarchical manner.
Formally, the fusing function is defined by
\begin{equation}
  \label{equ:equ2}
f_{l}(X)=
\left\{
\begin{aligned}
&h([g_{l}(X_O), f_{l+1}(X), g^{*}_{l}(X^{k}_R)]),l<L\\
&h([g_{l}(X_O), g^{*}_{l}(X^{k}_R)]), l = L
\end{aligned}
\right.
\end{equation}
where $h$ denotes the integration operator, which is a $1\times 1$ convolutional layer followed by a deconvolutional layer to ensure the same resolution.
$[\cdot]$ is the concatenation operator in channel-wise.
$g_{l}$ and $g^{*}_{l}$ are the reflection features of the $l$-th convolutional layer in the two sibling branches, respectively.

In the end, we add a convolutional layer with two filters for the saliency map prediction.
The numbers in Fig.~\ref{fig:framework} illustrate the detailed filter setting in each convolutional layer.
\subsection{Weighted Structural Loss}
Given the SOD training dataset $ S=\{(X_n,Y_n)\}^{N}_{n=1}$ with $N$ training pairs, where $X_n =
\{x^n_i,i = 1,...,T\}$ and $Y_n = \{y^n_i,i = 1,...,T\}$ are the input image and the binary ground-truth image with $T$ pixels, respectively.
$y^n_i = 1$ denotes the foreground pixel and $y^n_i = 0$ denotes the background pixel.
For notional simplicity, we subsequently drop the subscript $n$ and consider each image independently.
In most of existing SOD methods, the loss function used to train the network is the standard pixel-wise binary cross-entropy (BCE) loss:
\begin{equation}
  \label{equ:equ5}
\begin{aligned}
  \mathcal{L}_{bce}= -\sum_{i\in Y_{+}} \text{log~Pr}(y_{i}=1|X;\theta)\\-\sum_{i\in Y_{-}} \text{log~Pr}(y_{i}=0|X;\theta).
\end{aligned}
\end{equation}
where $\theta$ is the parameter of the network.
Pr$(y_i =1|X;\theta)\in [0,1]$ is the confidence score of the network prediction that measures how likely the pixel belong to the foreground.

However, for a typical natural image, the class distribution of salient/non-salient pixels is heavily imbalanced: most of the pixels in the ground truth are non-salient.
To automatically balance the loss between positive/negative classes, we introduce a class-balancing weight $\beta$ on a per-pixel term basis, following~\cite{xie2015holistically}.
Specifically, we define the following weighted cross-entropy loss function,
\begin{equation}
  \label{equ:equ4}
\begin{aligned}
  \mathcal{L}_{wbce}= - \beta \sum_{i\in Y_{+}} \text{log~Pr}(y_{i}=1|X;\theta)\\
  -(1-\beta)\sum_{i\in Y_{-}} \text{log~Pr}(y_{i}=0|X;\theta).
\end{aligned}
\end{equation}
The loss weight $\beta = |Y_{+}|/|Y|$, and $|Y_{+}|$ and $|Y_{-}|$ denote the foreground and background pixel number, respectively.

For saliency detection, it is also crucial to preserve the overall spatial structure and semantic content.
Thus, rather than only encouraging the pixels of the output prediction to match with the ground-truth using the above pixel-wise loss, we also minimize the differences between their multi-level features by a deep convolutional network~\cite{johnson2016perceptual}.
The main intuition behind this operator is that minimizing the difference between multi-level features, which encode low-level fine details and high-level coarse semantics, helps to
retain the spatial structure and semantic content of predictions.
Formally, let $\phi_{l}$ denotes the output of the $l$-th convolutional layer in a CNN, our semantic content (SC) loss is defined as
%
\begin{align}
\mathcal{L}_{sc} = \sum_{l=1}^{L}\lambda_{l}||\phi_{l}(Y;w)-\phi_{l}(\hat{Y};w)||_2,
  \label{equ:equ1}
\end{align}
where $\hat{Y}$ is the overall prediction, $w$ is the parameter of a pre-trained CNN and $\lambda_{l}$ is the trade-off parameter, controlling the influence of the loss in the $l$-th layer.
In our case, we use the light CNN-9 model~\cite{wu2015lightened} to calculate the above loss between the ground-truth and the prediction.
%
%

To overcome the blurry boundary problem~\cite{li2016deepsaliency}, we also introduce the smooth $L_1$ loss which encourages to keep the details of boundaries of salient objects.
Specifically, the smooth $L_1$ loss function is defined as
%
\begin{equation}
  \label{equ:equ2}
\mathcal{L}_{s1}=
\left\{
\begin{aligned}
&\frac{1}{2}||D||^2_2,&||D||_1<\epsilon\\
&\epsilon||D||_1-\frac{1}{2}\epsilon^2, &otherwise
\end{aligned}
\right.
\end{equation}
where $D=Y-\hat{Y}$ and $\epsilon$ is a predefined threshold. 
Following the practice in~\cite{xiao2018interactive}, we set $\epsilon=0.5$.
%
This training loss also helps to minimize pixel-level differences between the overall prediction and the ground-truth.

By taking all above loss functions together, we define our final loss function as
\begin{align}
\mathcal{L} = \text{arg min}~\mathcal{L}_{wbce}+\mu \mathcal{L}_{sc}+\gamma \mathcal{L}_{s1},
  \label{equ:equ1}
\end{align}
where $\mu$ and $\gamma$ are hyperparameters to balance the specific terms.
All the above losses are continuously differentiable, so we can use the standard stochastic gradient descent (SGD) method to obtain the optimal parameters.
In addition, we use $\lambda_{l}=1$, $\mu=0.01$ and $\gamma =20$ to optimize the final loss function for our experiments without further tuning.
\section{Experimental Results}
\subsection{Datasets and Evaluation Metrics}
To train our model, we adopt the \textbf{MSRA10K}~\cite{borji2015salient} dataset, which has 10,000 training images with high quality pixel-wise saliency annotations.
Most of images in this dataset have a single salient object.
To combat overfitting, we augment this dataset by random cropping and mirror reflection, producing 120,000 training images totally.

For the performance evaluation, we adopt seven public saliency detection datasets as follows:
\textbf{DUT-OMRON}~\cite{yang2013saliency} dataset has 5,168 high quality natural images.
Each image in this dataset has one or more objects with relatively complex image background.
\textbf{DUTS-TE} dataset is the test set of currently largest saliency detection benchmark (DUTS)~\cite{wang2017learning}.
It contains 5,019 images with high quality pixel-wise annotations.
\textbf{ECSSD}~\cite{shi2016hierarchical} dataset contains 1,000 natural images, in which many semantically meaningful and complex structures are included.
\textbf{HKU-IS-TE}~\cite{li2015visual} dataset has 1,447 images with pixel-wise annotations.
Images of this dataset are well chosen to include multiple disconnected objects or objects touching the image boundary.
\textbf{PASCAL-S}~\cite{li2014secrets} dataset is generated from the PASCAL VOC~\cite{Everingham2010ThePV} dataset and contains 850 natural images with segmentation-based masks.
\textbf{SED}~\cite{borj2015salient} dataset has two non-overlapped subsets, \emph{i.e.}, SED1 and SED2.
SED1 has 100 images each containing only one salient object, while SED2 has 100 images each containing two salient objects.
\textbf{SOD}~\cite{jiang2013salient} dataset has 300 images, in which many images contain multiple objects either with low contrast or touching the image boundary.

To evaluate the performance of varied SOD algorithms, we adopt four metrics, including the widely used precision-recall (PR) curves, F-measure, mean absolute error (MAE)~\cite{borji2015salient} and recently proposed S-measure~\cite{fan2017structure}.
%
The PR curve of a specific dataset exhibits the mean precision and recall of saliency maps at different thresholds.
The F-measure is a weighted mean of average precision and average recall, calculated by
\begin{equation}
  F_{\eta} =\frac{(1+\eta^2)\times Precision\times Recall}{\eta^2\times Precision \times Recall}.
    \label{equ:equ11}
\end{equation}
%
We set $\eta^2$ to be 0.3 to weigh precision more than recall as suggested in~\cite{borji2015salient}.

%
%
For fair comparison on non-salient regions, we also calculate the mean absolute error (MAE) by
\begin{equation}
MAE = \frac{1}{W\times H}\sum_{x=1}^{W}\sum_{y=1}^{H}|S(x,y)-G(x,y)|,
  \label{equ:equ12}
\end{equation}
where $W$ and $H$ are the width and height of the input image. $S(x,y)$ and $G(x,y)$ are the pixel values of the saliency map and the binary ground truth at $(x,y)$, respectively.

To evaluate the spatial structure similarities of saliency maps, we also calculate the S-measure, defined as
\begin{equation}
S_{\lambda} = \lambda*S_{o}+(1-\lambda)*S_{r},
  \label{equ:equ13}
\end{equation}
where $\lambda \in [0,1]$ is the balance parameter. $S_{o}$ and $S_{r}$ are the object-aware and region-aware structural similarity, respectively. We set $\lambda=0.5$ as suggested in~\cite{fan2017structure}.
\subsection{Implementation Details}
We implement our proposed model based on the Caffe toolbox~\cite{jia2014caffe} with the MATLAB 2016 platform.
We train and test our method in a quad-core PC machine with an NVIDIA Titan 1070 GPU (with 8G memory) and an i5-6600 CPU.
We perform training with the augmented training images from the MSRA10K dataset.
Following~\cite{zhang2017amulet,zhang2017learning}, we do not use validation set and train the model until its training loss converges.
The input image is uniformly resized into $384\times384\times3$ pixels and subtracted the ImageNet mean~\cite{deng2009imagenet}.
The weights of sibling branches are initialized from the VGG-16 model.
For the fusing branch, we initialize the weights by the ``msra'' method.
During the training, we use standard SGD method with batch size 12, momentum 0.9 and weight decay 0.0005.
We set the base learning rate to 1e-8 and decrease the learning rate by 10\% when training loss reaches a flat.
The training process converges after 150k iterations.
When testing, our proposed SOD algorithm runs at about \textbf{12 fps}.
%
The source code is publicly available at \textcolor[rgb]{1,0,0}{http://ice.dlut.edu.cn/lu/}.
\subsection{Comparison with the State-of-the-arts}
\begin{figure*}
\begin{center}
\begin{tabular}{@{}c@{}c@{}c@{}c}
\includegraphics[width=0.24\linewidth,height=3.17cm]{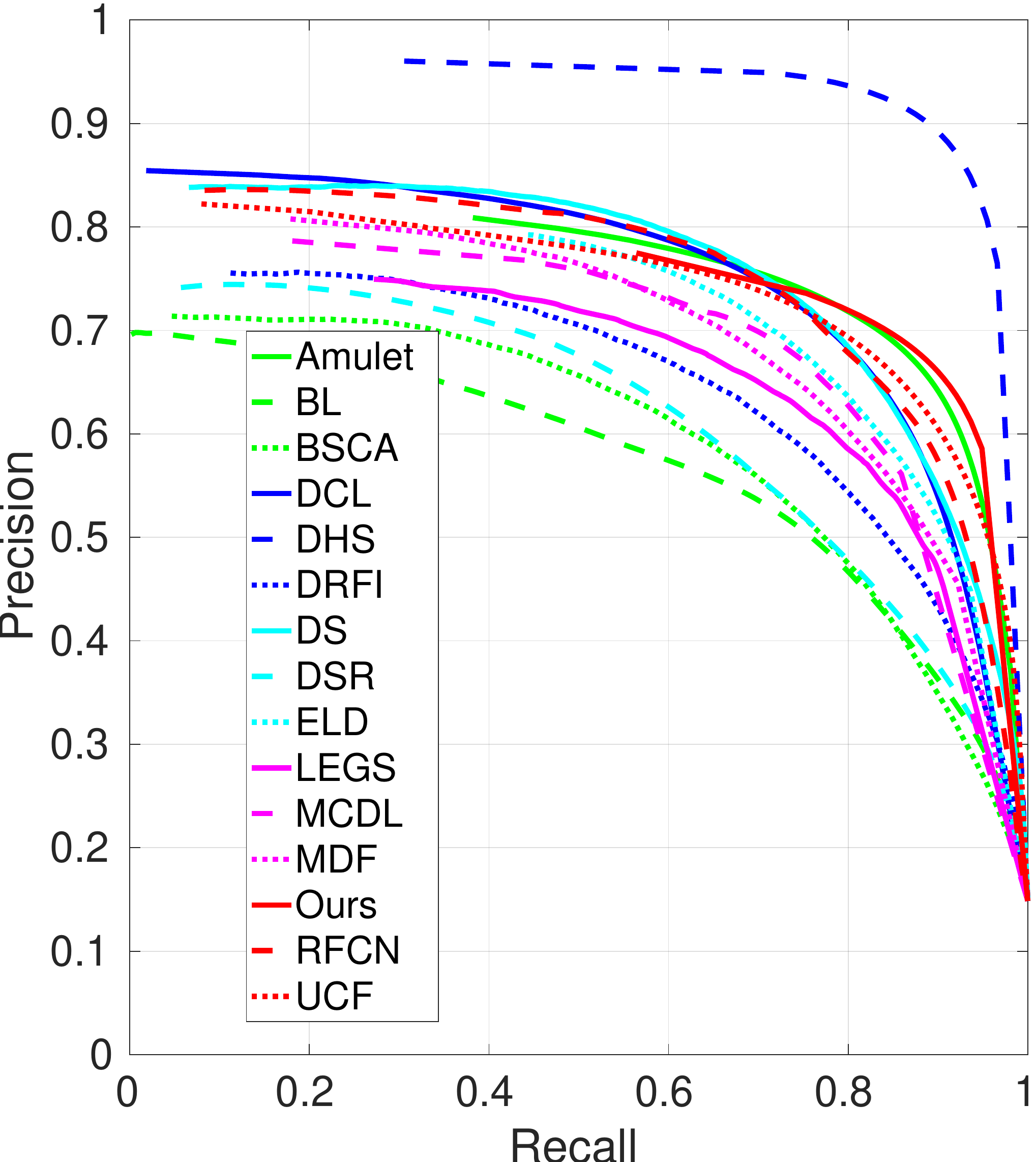} \ &
\includegraphics[width=0.24\linewidth,height=3.17cm]{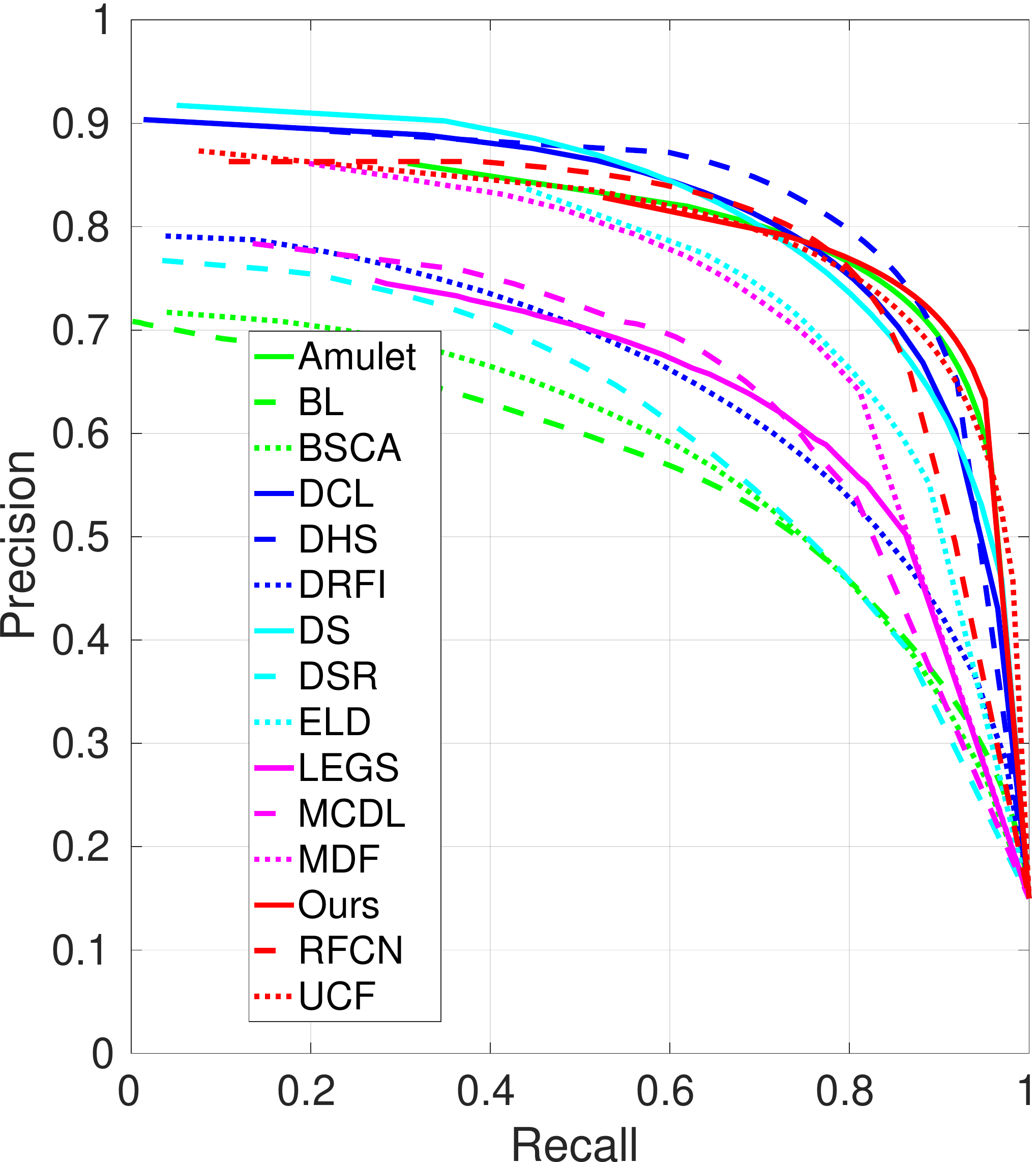} \ &
\includegraphics[width=0.24\linewidth,height=3.17cm]{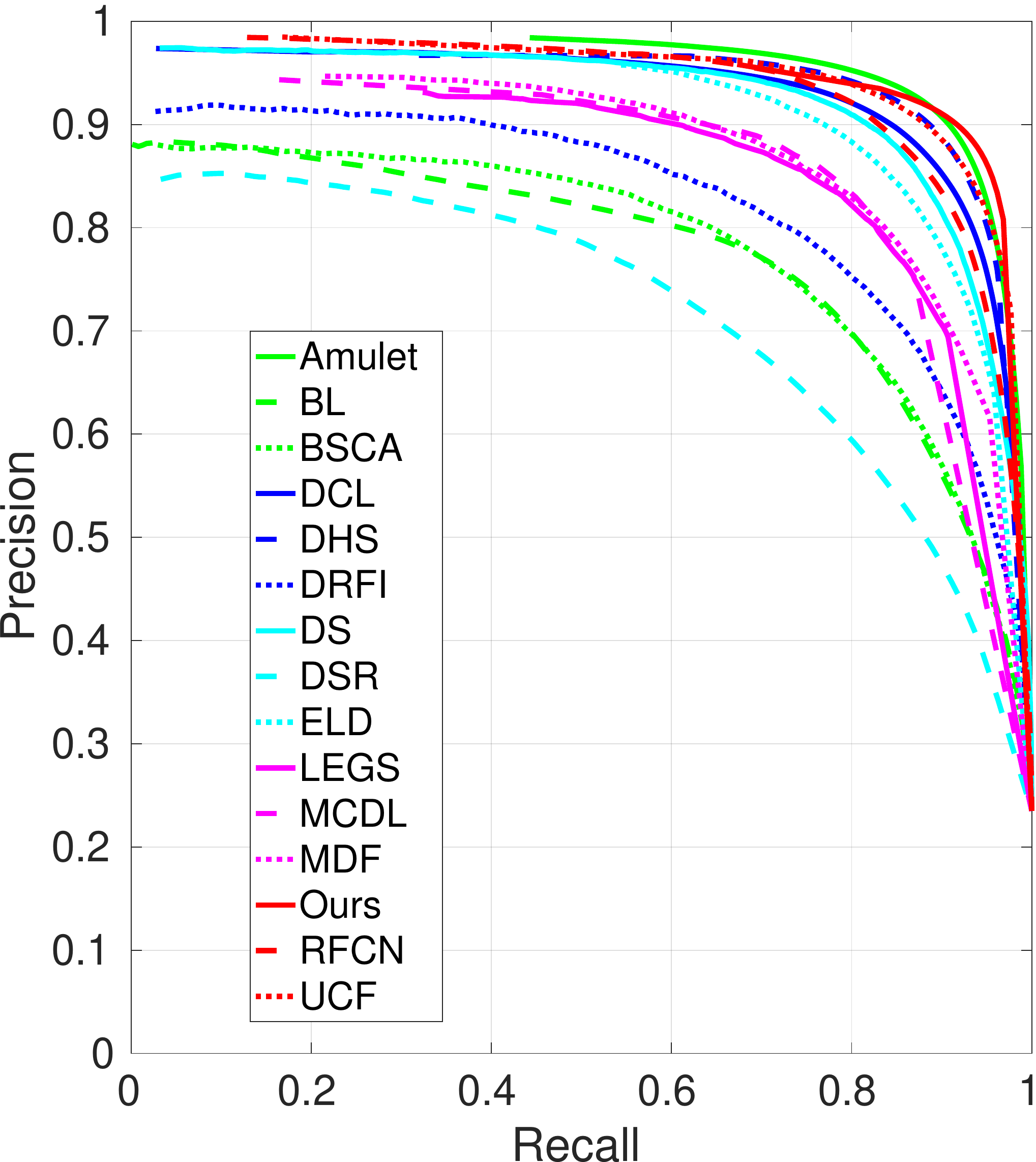} \ &
\includegraphics[width=0.24\linewidth,height=3.17cm]{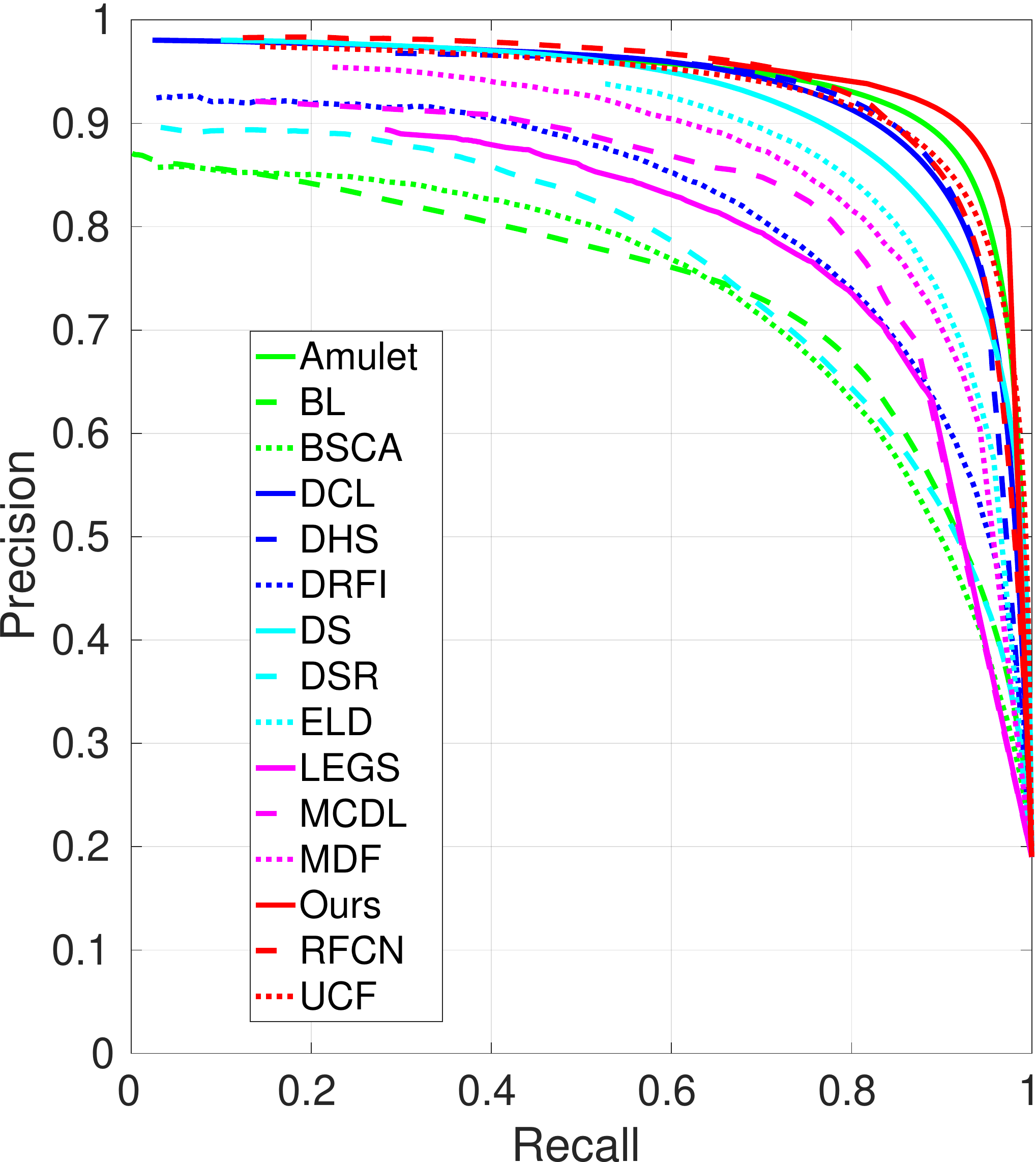} \ \\
 {\small(a) \textbf{DUT-OMRON}} & {\small(b) \textbf{DUTS-TE}} & {\small(c) \textbf{ECSSD}} & {\small(d) \textbf{HKU-IS-TE}}\\
\includegraphics[width=0.24\linewidth,height=3.17cm]{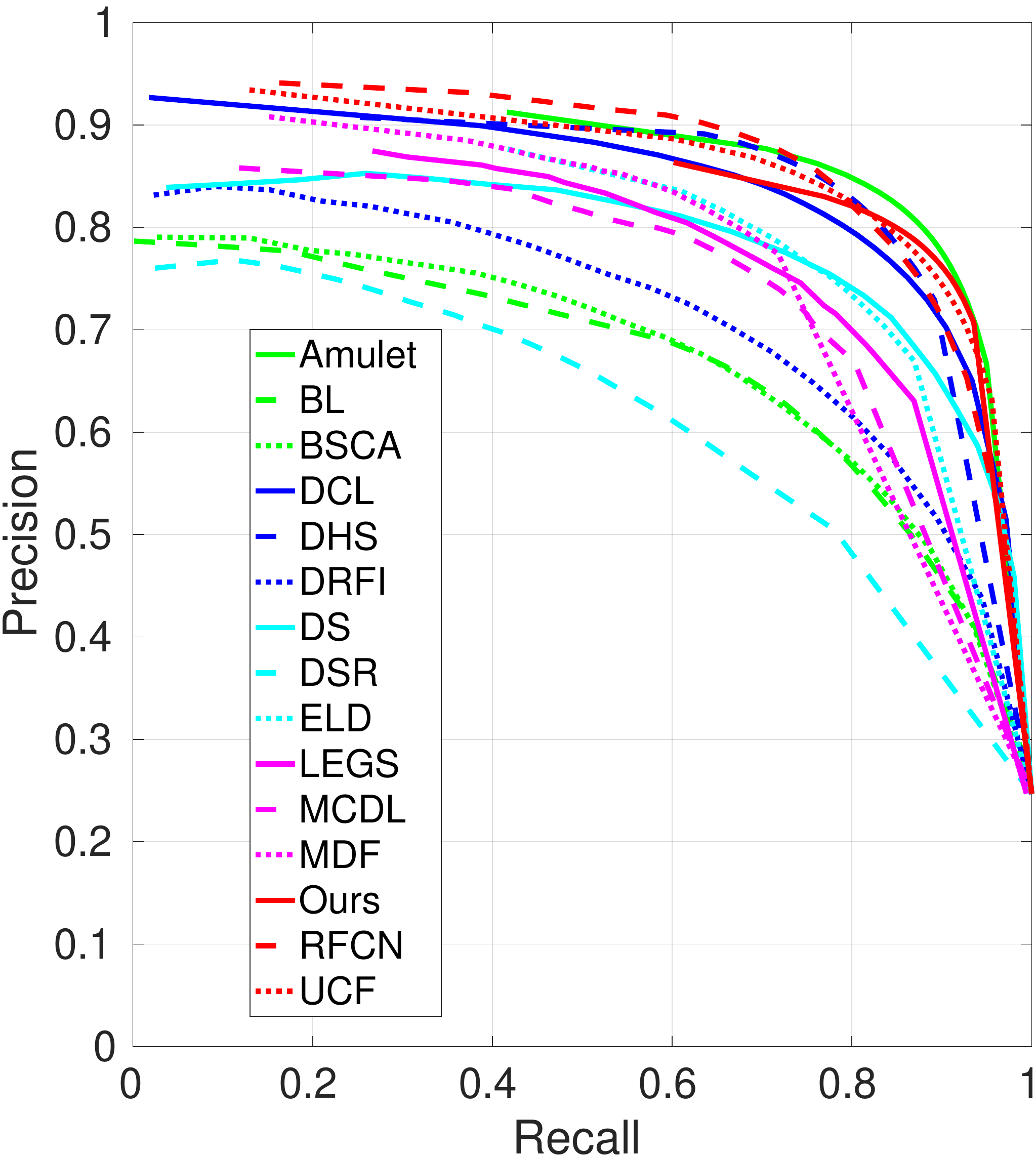} \ &
\includegraphics[width=0.24\linewidth,height=3.17cm]{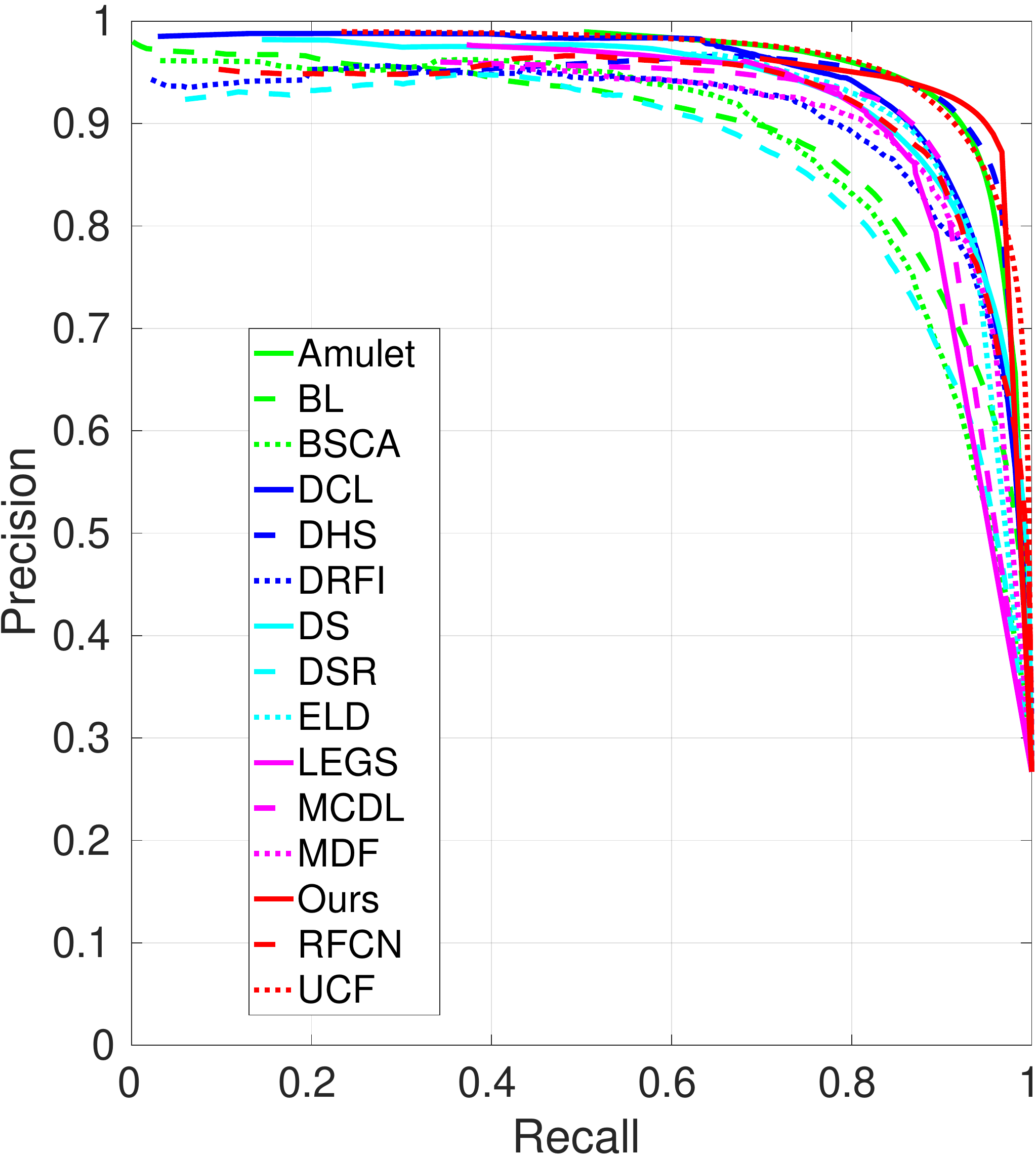} \ &
\includegraphics[width=0.24\linewidth,height=3.17cm]{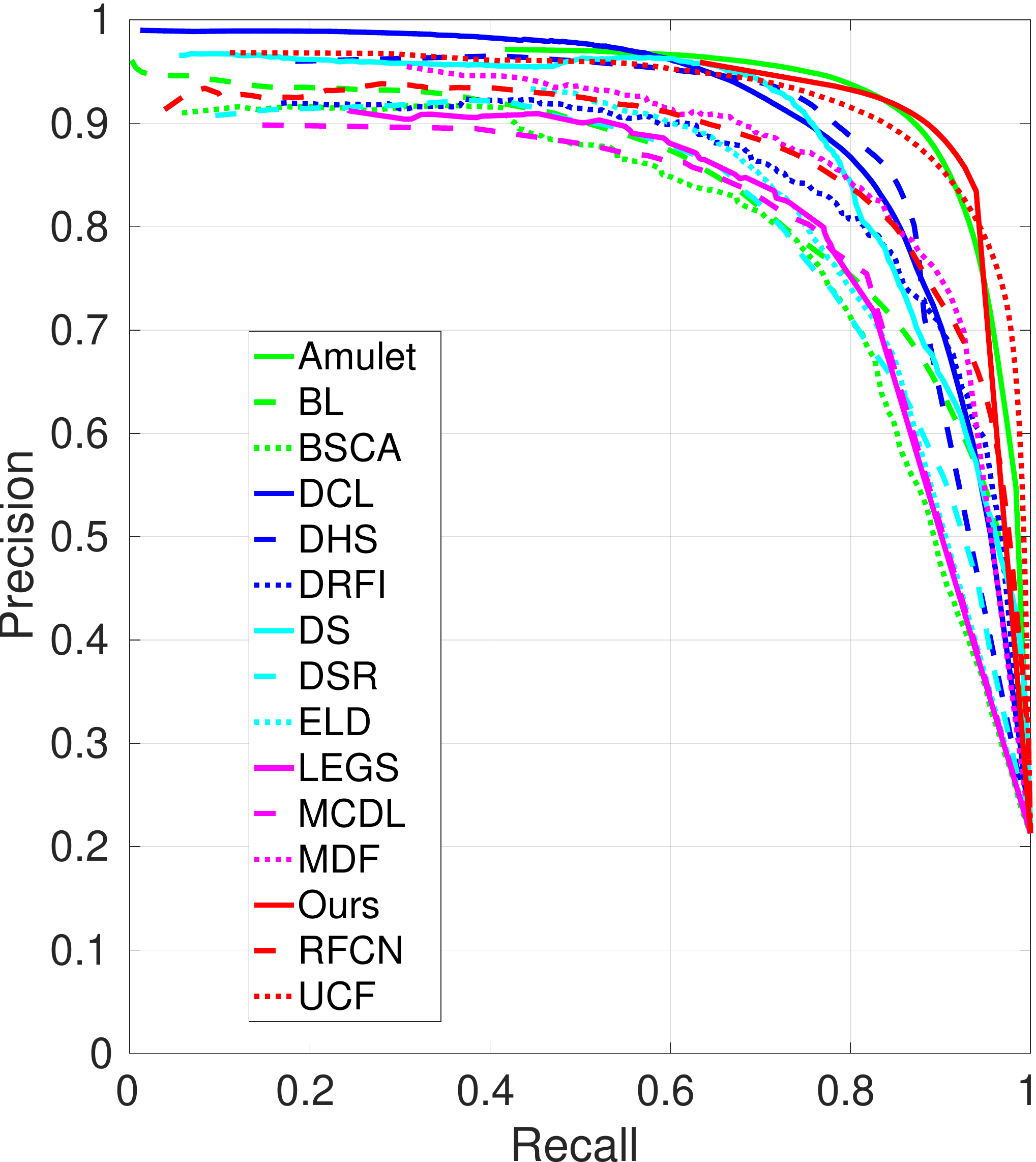} \ &
\includegraphics[width=0.24\linewidth,height=3.17cm]{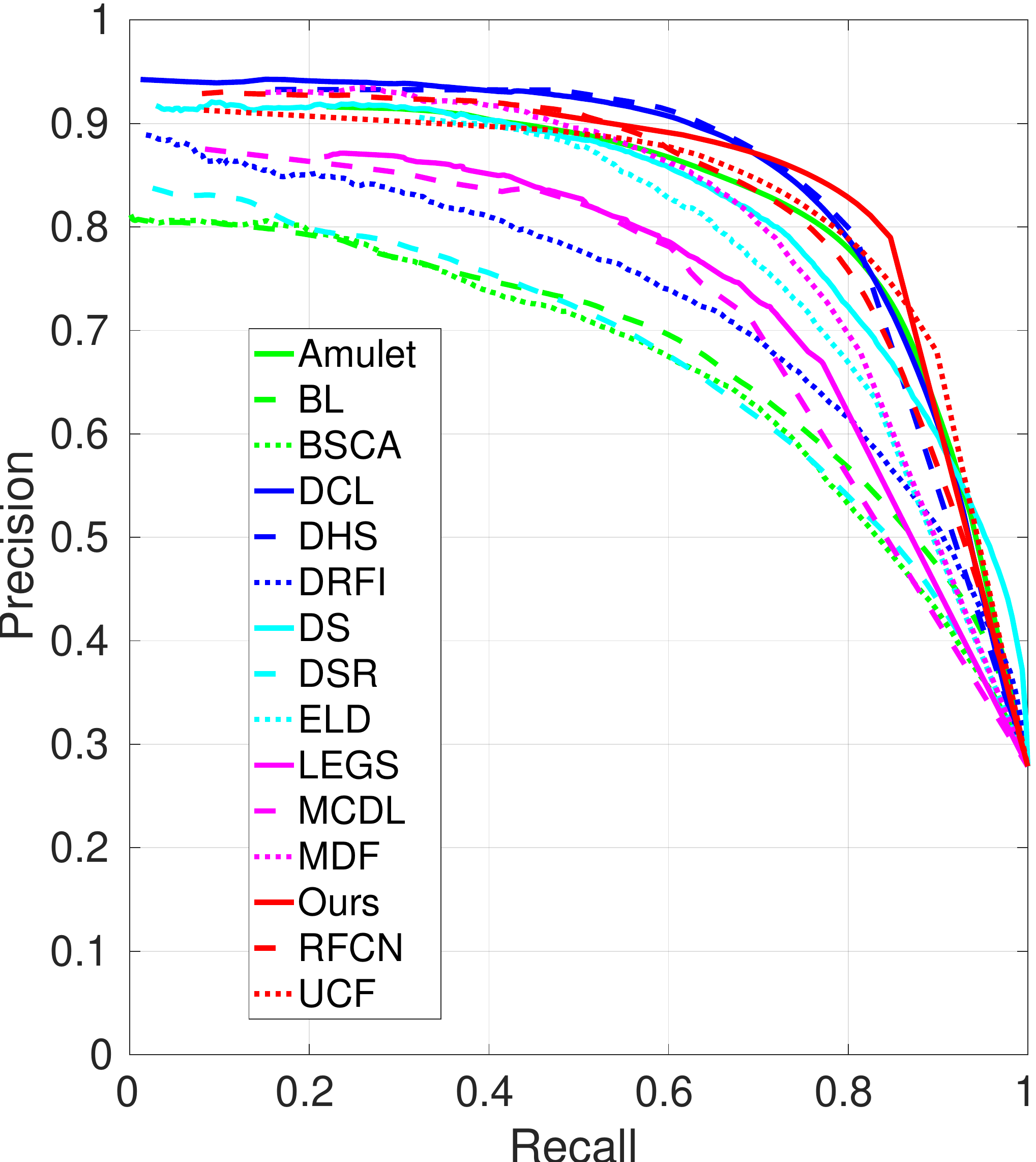} \ \\
 {\small(e) \textbf{PASCAL-S}} & {\small(f) \textbf{SED1}} & {\small(g) \textbf{SED2}} & {\small(h) \textbf{SOD}}\\
\\
\end{tabular}
\vspace{-6mm}
\caption{The PR curves of the proposed algorithm and other state-of-the-art methods.
\label{fig:PR-curve}}
\end{center}
\vspace{-5mm}
\end{figure*}

To fully evaluate the detection performance, we compare our proposed method with other 14 state-of-the-art ones, including 10 deep learning based algorithms (\textbf{Amulet}~\cite{zhang2017amulet}, \textbf{DCL}~\cite{li2016deep}, \textbf{DHS}~\cite{liu2016dhsnet}, \textbf{DS}~\cite{li2016deepsaliency}, \textbf{ELD}~\cite{lee2016deep}, \textbf{LEGS}~\cite{wang2015deep}, \textbf{MCDL}~\cite{zhao2015saliency}, \textbf{MDF}~\cite{li2015visual}, \textbf{RFCN}~\cite{wang2016saliency}, \textbf{UCF}~\cite{zhang2017learning}) and 4 conventional algorithms (\textbf{BL}~\cite{tong2015salient}, \textbf{BSCA}~\cite{qin2015saliency}, \textbf{DRFI}~\cite{jiang2013salient}, \textbf{DSR}~\cite{li2013saliency}).
%
For fair comparison, we use either the implementations with recommended parameter settings or the saliency maps provided by the authors.
\setlength{\tabcolsep}{4pt}
\begin{table*}
\begin{center}
\doublerulesep=0.4pt
\begin{tabular}{|c|c|c|c|c|c|c|c|c|c|c|c|c|c|c|c|c|c|c|c|c|c|c|c|c|||c|c|c|c|c|c|c|c|||}
\hline
\multicolumn{4}{|c|}{}
&\multicolumn{6}{|c|}{\textbf{DUT-OMRON}}
&\multicolumn{6}{|c|}{\textbf{DUTS-TE}}
&\multicolumn{6}{|c|}{\textbf{ECSSD}}
&\multicolumn{6}{|c|}{\textbf{HKU-IS-TE}}
\\
\hline
\hline
\multicolumn{4}{|c|}{Methods}
&\multicolumn{2}{|c|}{$F_\eta$}&\multicolumn{2}{|c|}{$MAE$}&\multicolumn{2}{|c|}{$S_\lambda$}
&\multicolumn{2}{|c|}{$F_\eta$}&\multicolumn{2}{|c|}{$MAE$}&\multicolumn{2}{|c|}{$S_\lambda$}
&\multicolumn{2}{|c|}{$F_\eta$}&\multicolumn{2}{|c|}{$MAE$}&\multicolumn{2}{|c|}{$S_\lambda$}
&\multicolumn{2}{|c|}{$F_\eta$}&\multicolumn{2}{|c|}{$MAE$}&\multicolumn{2}{|c|}{$S_\lambda$}
\\
\hline
\hline
\multicolumn{4}{|c|}{\textbf{Ours}}
&\multicolumn{2}{|c|}{\textcolor[rgb]{1,0,0}{0.696}}&\multicolumn{2}{|c|}{\textcolor[rgb]{1,0,0}{0.086}}&\multicolumn{2}{|c|}{\textcolor[rgb]{1,0,0}{0.774}}
&\multicolumn{2}{|c|}{\textcolor[rgb]{0,1,0}{0.716}}&\multicolumn{2}{|c|}{\textcolor[rgb]{0,1,0}{0.083}}&\multicolumn{2}{|c|}{\textcolor[rgb]{0,1,0}{0.799}}
&\multicolumn{2}{|c|}{\textcolor[rgb]{1,0,0}{0.880}}&\multicolumn{2}{|c|}{\textcolor[rgb]{1,0,0}{0.052}}&\multicolumn{2}{|c|}{\textcolor[rgb]{1,0,0}{0.897}}
&\multicolumn{2}{|c|}{\textcolor[rgb]{1,0,0}{0.875}}&\multicolumn{2}{|c|}{\textcolor[rgb]{1,0,0}{0.040}}&\multicolumn{2}{|c|}{\textcolor[rgb]{1,0,0}{0.905}}
\\
\multicolumn{4}{|c|}{\textbf{Amulet}~\cite{zhang2017amulet}}
&\multicolumn{2}{|c|}{\textcolor[rgb]{0,0,1}{0.647}}&\multicolumn{2}{|c|}{0.098}&\multicolumn{2}{|c|}{\textcolor[rgb]{0,1,0}{0.771}}
&\multicolumn{2}{|c|}{0.682}&\multicolumn{2}{|c|}{\textcolor[rgb]{0,0,1}{0.085}}&\multicolumn{2}{|c|}{\textcolor[rgb]{0,0,1}{0.796}}
&\multicolumn{2}{|c|}{\textcolor[rgb]{0,0,1}{0.868}}&\multicolumn{2}{|c|}{\textcolor[rgb]{0,1,0}{0.059}}&\multicolumn{2}{|c|}{\textcolor[rgb]{0,1,0}{0.894}}
&\multicolumn{2}{|c|}{0.843}&\multicolumn{2}{|c|}{\textcolor[rgb]{0,1,0}{0.050}}&\multicolumn{2}{|c|}{\textcolor[rgb]{0,1,0}{0.886}}
\\
\multicolumn{4}{|c|}{\textbf{DCL}~\cite{li2016deep}}
&\multicolumn{2}{|c|}{\textcolor[rgb]{0,1,0}{0.684}}&\multicolumn{2}{|c|}{0.157}&\multicolumn{2}{|c|}{0.743}
&\multicolumn{2}{|c|}{\textcolor[rgb]{0,0,1}{0.714}}&\multicolumn{2}{|c|}{0.150}&\multicolumn{2}{|c|}{0.785}
&\multicolumn{2}{|c|}{0.829}&\multicolumn{2}{|c|}{0.149}&\multicolumn{2}{|c|}{0.863}
&\multicolumn{2}{|c|}{\textcolor[rgb]{0,0,1}{0.853}}&\multicolumn{2}{|c|}{0.136}&\multicolumn{2}{|c|}{0.859}
\\
\multicolumn{4}{|c|}{\textbf{DHS}~\cite{liu2016dhsnet}}
&\multicolumn{2}{|c|}{--}&\multicolumn{2}{|c|}{--}&\multicolumn{2}{|c|}{--}
&\multicolumn{2}{|c|}{\textcolor[rgb]{1,0,0}{0.724}}&\multicolumn{2}{|c|}{\textcolor[rgb]{1,0,0}{0.066}}&\multicolumn{2}{|c|}{\textcolor[rgb]{1,0,0}{0.809}}
&\multicolumn{2}{|c|}{\textcolor[rgb]{0,1,0}{0.872}}&\multicolumn{2}{|c|}{\textcolor[rgb]{0,0,1}{0.060}}&\multicolumn{2}{|c|}{0.884}
&\multicolumn{2}{|c|}{\textcolor[rgb]{0,1,0}{0.854}}&\multicolumn{2}{|c|}{\textcolor[rgb]{0,0,1}{0.053}}&\multicolumn{2}{|c|}{0.869}
\\
\multicolumn{4}{|c|}{\textbf{DS}~\cite{li2016deepsaliency}}
&\multicolumn{2}{|c|}{0.603}&\multicolumn{2}{|c|}{0.120}&\multicolumn{2}{|c|}{0.741}
&\multicolumn{2}{|c|}{0.632}&\multicolumn{2}{|c|}{0.091}&\multicolumn{2}{|c|}{0.790}
&\multicolumn{2}{|c|}{0.826}&\multicolumn{2}{|c|}{0.122}&\multicolumn{2}{|c|}{0.821}
&\multicolumn{2}{|c|}{0.787}&\multicolumn{2}{|c|}{0.077}&\multicolumn{2}{|c|}{0.854}
\\
\multicolumn{4}{|c|}{\textbf{ELD}~\cite{lee2016deep}}
&\multicolumn{2}{|c|}{0.611}&\multicolumn{2}{|c|}{0.092}&\multicolumn{2}{|c|}{0.743}
&\multicolumn{2}{|c|}{0.628}&\multicolumn{2}{|c|}{0.098}&\multicolumn{2}{|c|}{0.749}
&\multicolumn{2}{|c|}{0.810}&\multicolumn{2}{|c|}{0.080}&\multicolumn{2}{|c|}{0.839}
&\multicolumn{2}{|c|}{0.776}&\multicolumn{2}{|c|}{0.072}&\multicolumn{2}{|c|}{0.823}
\\
\multicolumn{4}{|c|}{\textbf{LEGS}~\cite{wang2015deep}}
&\multicolumn{2}{|c|}{0.592}&\multicolumn{2}{|c|}{0.133}&\multicolumn{2}{|c|}{0.701}
&\multicolumn{2}{|c|}{0.585}&\multicolumn{2}{|c|}{0.138}&\multicolumn{2}{|c|}{0.687}
&\multicolumn{2}{|c|}{0.785}&\multicolumn{2}{|c|}{0.118}&\multicolumn{2}{|c|}{0.787}
&\multicolumn{2}{|c|}{0.732}&\multicolumn{2}{|c|}{0.118}&\multicolumn{2}{|c|}{0.745}
\\
\multicolumn{4}{|c|}{\textbf{MCDL}~\cite{zhao2015saliency}}
&\multicolumn{2}{|c|}{0.625}&\multicolumn{2}{|c|}{\textcolor[rgb]{0,1,0}{0.089}}&\multicolumn{2}{|c|}{0.739}
&\multicolumn{2}{|c|}{0.594}&\multicolumn{2}{|c|}{0.105}&\multicolumn{2}{|c|}{0.706}
&\multicolumn{2}{|c|}{0.796}&\multicolumn{2}{|c|}{0.101}&\multicolumn{2}{|c|}{0.803}
&\multicolumn{2}{|c|}{0.760}&\multicolumn{2}{|c|}{0.091}&\multicolumn{2}{|c|}{0.786}
\\
\multicolumn{4}{|c|}{\textbf{MDF}~\cite{li2015visual}}
&\multicolumn{2}{|c|}{0.644}&\multicolumn{2}{|c|}{\textcolor[rgb]{0,0,1}{0.092}}&\multicolumn{2}{|c|}{0.703}
&\multicolumn{2}{|c|}{0.673}&\multicolumn{2}{|c|}{0.100}&\multicolumn{2}{|c|}{0.723}
&\multicolumn{2}{|c|}{0.807}&\multicolumn{2}{|c|}{0.105}&\multicolumn{2}{|c|}{0.776}
&\multicolumn{2}{|c|}{0.802}&\multicolumn{2}{|c|}{0.095}&\multicolumn{2}{|c|}{0.779}
\\
\multicolumn{4}{|c|}{\textbf{RFCN}~\cite{wang2016saliency}}
&\multicolumn{2}{|c|}{0.627}&\multicolumn{2}{|c|}{0.111}&\multicolumn{2}{|c|}{\textcolor[rgb]{0,0,1}{0.752}}
&\multicolumn{2}{|c|}{0.712}&\multicolumn{2}{|c|}{0.090}&\multicolumn{2}{|c|}{0.784}
&\multicolumn{2}{|c|}{0.834}&\multicolumn{2}{|c|}{0.107}&\multicolumn{2}{|c|}{0.852}
&\multicolumn{2}{|c|}{0.838}&\multicolumn{2}{|c|}{0.088}&\multicolumn{2}{|c|}{0.860}
\\
\multicolumn{4}{|c|}{\textbf{UCF}~\cite{zhang2017learning}}
&\multicolumn{2}{|c|}{0.621}&\multicolumn{2}{|c|}{0.120}&\multicolumn{2}{|c|}{0.748}
&\multicolumn{2}{|c|}{0.635}&\multicolumn{2}{|c|}{0.112}&\multicolumn{2}{|c|}{0.777}
&\multicolumn{2}{|c|}{0.844}&\multicolumn{2}{|c|}{0.069}&\multicolumn{2}{|c|}{\textcolor[rgb]{0,0,1}{0.884}}
&\multicolumn{2}{|c|}{0.823}&\multicolumn{2}{|c|}{0.061}&\multicolumn{2}{|c|}{\textcolor[rgb]{0,0,1}{0.874}}
\\
\hline
\hline
\multicolumn{4}{|c|}{\textbf{BL}~\cite{tong2015salient}}
&\multicolumn{2}{|c|}{0.499}&\multicolumn{2}{|c|}{0.239}&\multicolumn{2}{|c|}{0.625}
&\multicolumn{2}{|c|}{0.490}&\multicolumn{2}{|c|}{0.238}&\multicolumn{2}{|c|}{0.615}
&\multicolumn{2}{|c|}{0.684}&\multicolumn{2}{|c|}{0.216}&\multicolumn{2}{|c|}{0.714}
&\multicolumn{2}{|c|}{0.666}&\multicolumn{2}{|c|}{0.207}&\multicolumn{2}{|c|}{0.702}
\\
\multicolumn{4}{|c|}{\textbf{BSCA}~\cite{qin2015saliency}}
&\multicolumn{2}{|c|}{0.509}&\multicolumn{2}{|c|}{0.190}&\multicolumn{2}{|c|}{0.652}
&\multicolumn{2}{|c|}{0.500}&\multicolumn{2}{|c|}{0.196}&\multicolumn{2}{|c|}{0.633}
&\multicolumn{2}{|c|}{0.705}&\multicolumn{2}{|c|}{0.182}&\multicolumn{2}{|c|}{0.725}
&\multicolumn{2}{|c|}{0.658}&\multicolumn{2}{|c|}{0.175}&\multicolumn{2}{|c|}{0.705}
\\
\multicolumn{4}{|c|}{\textbf{DRFI}~\cite{jiang2013salient}}
&\multicolumn{2}{|c|}{0.550}&\multicolumn{2}{|c|}{0.138}&\multicolumn{2}{|c|}{0.688}
&\multicolumn{2}{|c|}{0.541}&\multicolumn{2}{|c|}{0.175}&\multicolumn{2}{|c|}{0.662}
&\multicolumn{2}{|c|}{0.733}&\multicolumn{2}{|c|}{0.164}&\multicolumn{2}{|c|}{0.752}
&\multicolumn{2}{|c|}{0.726}&\multicolumn{2}{|c|}{0.145}&\multicolumn{2}{|c|}{0.743}
\\
\multicolumn{4}{|c|}{\textbf{DSR}~\cite{li2013saliency}}
&\multicolumn{2}{|c|}{0.524}&\multicolumn{2}{|c|}{0.139}&\multicolumn{2}{|c|}{0.660}
&\multicolumn{2}{|c|}{0.518}&\multicolumn{2}{|c|}{0.145}&\multicolumn{2}{|c|}{0.646}
&\multicolumn{2}{|c|}{0.662}&\multicolumn{2}{|c|}{0.178}&\multicolumn{2}{|c|}{0.731}
&\multicolumn{2}{|c|}{0.682}&\multicolumn{2}{|c|}{0.142}&\multicolumn{2}{|c|}{0.701}
\\
\hline
\end{tabular}
\vspace{-2mm}
\caption{Quantitative comparison with 15 methods on 4 large-scale datasets. The best three results are shown in \textcolor[rgb]{1,0,0}{red},~\textcolor[rgb]{0,1,0}{green} and \textcolor[rgb]{0,0,1}{blue}, respectively. ``--'' means corresponding methods are trained on that dataset. Our method ranks first or second on these datasets. }
\label{table:fauc1}
\end{center}
\vspace{-4mm}
\end{table*}
\setlength{\tabcolsep}{4pt}
\begin{table*}
\begin{center}
\doublerulesep=0.4pt
\begin{tabular}{|c|c|c|c|c|c|c|c|c|c|c|c|c|c|c|c|c|c|c|c|c|c|c|c|c|||c|c|c|c|c|c|c|c|||}
\hline
\multicolumn{4}{|c|}{}
&\multicolumn{6}{|c|}{\textbf{PASCAL-S}}
&\multicolumn{6}{|c|}{\textbf{SED1}}
&\multicolumn{6}{|c|}{\textbf{SED2}}
&\multicolumn{6}{|c|}{\textbf{SOD}}
\\
\hline
\hline
\multicolumn{4}{|c|}{Methods}
&\multicolumn{2}{|c|}{$F_\eta$}&\multicolumn{2}{|c|}{$MAE$}&\multicolumn{2}{|c|}{$S_\lambda$}
&\multicolumn{2}{|c|}{$F_\eta$}&\multicolumn{2}{|c|}{$MAE$}&\multicolumn{2}{|c|}{$S_\lambda$}
&\multicolumn{2}{|c|}{$F_\eta$}&\multicolumn{2}{|c|}{$MAE$}&\multicolumn{2}{|c|}{$S_\lambda$}
&\multicolumn{2}{|c|}{$F_\eta$}&\multicolumn{2}{|c|}{$MAE$}&\multicolumn{2}{|c|}{$S_\lambda$}
\\
\hline
\hline
\multicolumn{4}{|c|}{\textbf{Ours}}
&\multicolumn{2}{|c|}{\textcolor[rgb]{0,1,0}{0.772}}&\multicolumn{2}{|c|}{\textcolor[rgb]{0,0,1}{0.104}}&\multicolumn{2}{|c|}{\textcolor[rgb]{0,1,0}{0.809}}
&\multicolumn{2}{|c|}{\textcolor[rgb]{1,0,0}{0.913}}&\multicolumn{2}{|c|}{\textcolor[rgb]{1,0,0}{0.048}}&\multicolumn{2}{|c|}{\textcolor[rgb]{1,0,0}{0.905}}
&\multicolumn{2}{|c|}{\textcolor[rgb]{1,0,0}{0.871}}&\multicolumn{2}{|c|}{\textcolor[rgb]{1,0,0}{0.048}}&\multicolumn{2}{|c|}{\textcolor[rgb]{1,0,0}{0.870}}
&\multicolumn{2}{|c|}{\textcolor[rgb]{1,0,0}{0.789}}&\multicolumn{2}{|c|}{\textcolor[rgb]{1,0,0}{0.123}}&\multicolumn{2}{|c|}{\textcolor[rgb]{1,0,0}{0.772}}
\\
\multicolumn{4}{|c|}{\textbf{Amulet}~\cite{zhang2017amulet}}
&\multicolumn{2}{|c|}{\textcolor[rgb]{0,0,1}{0.768}}&\multicolumn{2}{|c|}{\textcolor[rgb]{0,1,0}{0.098}}&\multicolumn{2}{|c|}{\textcolor[rgb]{1,0,0}{0.820}}
&\multicolumn{2}{|c|}{\textcolor[rgb]{0,1,0}{0.892}}&\multicolumn{2}{|c|}{\textcolor[rgb]{0,0,1}{0.060}}&\multicolumn{2}{|c|}{0.893}
&\multicolumn{2}{|c|}{\textcolor[rgb]{0,1,0}{0.830}}&\multicolumn{2}{|c|}{\textcolor[rgb]{0,1,0}{0.062}}&\multicolumn{2}{|c|}{\textcolor[rgb]{0,1,0}{0.852}}
&\multicolumn{2}{|c|}{\textcolor[rgb]{0,0,1}{0.745}}&\multicolumn{2}{|c|}{\textcolor[rgb]{0,0,1}{0.144}}&\multicolumn{2}{|c|}{\textcolor[rgb]{0,0,1}{0.753}}
\\
\multicolumn{4}{|c|}{\textbf{DCL}~\cite{li2016deep}}
&\multicolumn{2}{|c|}{0.714}&\multicolumn{2}{|c|}{0.181}&\multicolumn{2}{|c|}{0.791}
&\multicolumn{2}{|c|}{0.855}&\multicolumn{2}{|c|}{0.151}&\multicolumn{2}{|c|}{0.845}
&\multicolumn{2}{|c|}{0.795}&\multicolumn{2}{|c|}{0.157}&\multicolumn{2}{|c|}{0.760}
&\multicolumn{2}{|c|}{0.741}&\multicolumn{2}{|c|}{0.194}&\multicolumn{2}{|c|}{0.748}
\\
\multicolumn{4}{|c|}{\textbf{DHS}~\cite{liu2016dhsnet}}
&\multicolumn{2}{|c|}{\textcolor[rgb]{1,0,0}{0.777}}&\multicolumn{2}{|c|}{\textcolor[rgb]{1,0,0}{0.095}}&\multicolumn{2}{|c|}{\textcolor[rgb]{0,0,1}{0.807}}
&\multicolumn{2}{|c|}{\textcolor[rgb]{0,0,1}{0.888}}&\multicolumn{2}{|c|}{\textcolor[rgb]{0,1,0}{0.055}}&\multicolumn{2}{|c|}{\textcolor[rgb]{0,0,1}{0.894}}
&\multicolumn{2}{|c|}{\textcolor[rgb]{0,0,1}{0.822}}&\multicolumn{2}{|c|}{0.080}&\multicolumn{2}{|c|}{0.796}
&\multicolumn{2}{|c|}{\textcolor[rgb]{0,1,0}{0.775}}&\multicolumn{2}{|c|}{\textcolor[rgb]{0,1,0}{0.129}}&\multicolumn{2}{|c|}{0.750}
\\
\multicolumn{4}{|c|}{\textbf{DS}~\cite{li2016deepsaliency}}
&\multicolumn{2}{|c|}{0.659}&\multicolumn{2}{|c|}{0.176}&\multicolumn{2}{|c|}{0.739}
&\multicolumn{2}{|c|}{0.845}&\multicolumn{2}{|c|}{0.093}&\multicolumn{2}{|c|}{0.859}
&\multicolumn{2}{|c|}{0.754}&\multicolumn{2}{|c|}{0.123}&\multicolumn{2}{|c|}{0.776}
&\multicolumn{2}{|c|}{0.698}&\multicolumn{2}{|c|}{0.189}&\multicolumn{2}{|c|}{0.712}
\\
\multicolumn{4}{|c|}{\textbf{ELD}~\cite{lee2016deep}}
&\multicolumn{2}{|c|}{0.718}&\multicolumn{2}{|c|}{0.123}&\multicolumn{2}{|c|}{0.757}
&\multicolumn{2}{|c|}{0.872}&\multicolumn{2}{|c|}{0.067}&\multicolumn{2}{|c|}{0.864}
&\multicolumn{2}{|c|}{0.759}&\multicolumn{2}{|c|}{0.103}&\multicolumn{2}{|c|}{0.769}
&\multicolumn{2}{|c|}{0.712}&\multicolumn{2}{|c|}{0.155}&\multicolumn{2}{|c|}{0.705}
\\
\multicolumn{4}{|c|}{\textbf{LEGS}~\cite{wang2015deep}}
&\multicolumn{2}{|c|}{--}&\multicolumn{2}{|c|}{--}&\multicolumn{2}{|c|}{--}
&\multicolumn{2}{|c|}{0.854}&\multicolumn{2}{|c|}{0.103}&\multicolumn{2}{|c|}{0.828}
&\multicolumn{2}{|c|}{0.736}&\multicolumn{2}{|c|}{0.124}&\multicolumn{2}{|c|}{0.716}
&\multicolumn{2}{|c|}{0.683}&\multicolumn{2}{|c|}{0.196}&\multicolumn{2}{|c|}{0.657}
\\
\multicolumn{4}{|c|}{\textbf{MCDL}~\cite{zhao2015saliency}}
&\multicolumn{2}{|c|}{0.691}&\multicolumn{2}{|c|}{0.145}&\multicolumn{2}{|c|}{0.719}
&\multicolumn{2}{|c|}{0.878}&\multicolumn{2}{|c|}{0.077}&\multicolumn{2}{|c|}{0.855}
&\multicolumn{2}{|c|}{0.757}&\multicolumn{2}{|c|}{0.116}&\multicolumn{2}{|c|}{0.742}
&\multicolumn{2}{|c|}{0.677}&\multicolumn{2}{|c|}{0.181}&\multicolumn{2}{|c|}{0.650}
\\
\multicolumn{4}{|c|}{\textbf{MDF}~\cite{li2015visual}}
&\multicolumn{2}{|c|}{0.709}&\multicolumn{2}{|c|}{0.146}&\multicolumn{2}{|c|}{0.692}
&\multicolumn{2}{|c|}{0.842}&\multicolumn{2}{|c|}{0.099}&\multicolumn{2}{|c|}{0.833}
&\multicolumn{2}{|c|}{0.800}&\multicolumn{2}{|c|}{0.101}&\multicolumn{2}{|c|}{0.772}
&\multicolumn{2}{|c|}{0.721}&\multicolumn{2}{|c|}{0.165}&\multicolumn{2}{|c|}{0.674}
\\
\multicolumn{4}{|c|}{\textbf{RFCN}~\cite{wang2016saliency}}
&\multicolumn{2}{|c|}{0.751}&\multicolumn{2}{|c|}{0.132}&\multicolumn{2}{|c|}{0.799}
&\multicolumn{2}{|c|}{0.850}&\multicolumn{2}{|c|}{0.117}&\multicolumn{2}{|c|}{0.832}
&\multicolumn{2}{|c|}{0.767}&\multicolumn{2}{|c|}{0.113}&\multicolumn{2}{|c|}{0.784}
&\multicolumn{2}{|c|}{0.743}&\multicolumn{2}{|c|}{0.170}&\multicolumn{2}{|c|}{0.730}
\\
\multicolumn{4}{|c|}{\textbf{UCF}~\cite{zhang2017learning}}
&\multicolumn{2}{|c|}{0.735}&\multicolumn{2}{|c|}{0.115}&\multicolumn{2}{|c|}{0.806}
&\multicolumn{2}{|c|}{0.865}&\multicolumn{2}{|c|}{0.063}&\multicolumn{2}{|c|}{\textcolor[rgb]{0,1,0}{0.896}}
&\multicolumn{2}{|c|}{0.810}&\multicolumn{2}{|c|}{\textcolor[rgb]{0,0,1}{0.068}}&\multicolumn{2}{|c|}{\textcolor[rgb]{0,0,1}{0.846}}
&\multicolumn{2}{|c|}{0.738}&\multicolumn{2}{|c|}{0.148}&\multicolumn{2}{|c|}{\textcolor[rgb]{0,1,0}{0.762}}
\\
\hline
\hline
\multicolumn{4}{|c|}{\textbf{BL}~\cite{tong2015salient}}
&\multicolumn{2}{|c|}{0.574}&\multicolumn{2}{|c|}{0.249}&\multicolumn{2}{|c|}{0.647}
&\multicolumn{2}{|c|}{0.780}&\multicolumn{2}{|c|}{0.185}&\multicolumn{2}{|c|}{0.783}
&\multicolumn{2}{|c|}{0.713}&\multicolumn{2}{|c|}{0.186}&\multicolumn{2}{|c|}{0.705}
&\multicolumn{2}{|c|}{0.580}&\multicolumn{2}{|c|}{0.267}&\multicolumn{2}{|c|}{0.625}
\\
\multicolumn{4}{|c|}{\textbf{BSCA}~\cite{qin2015saliency}}
&\multicolumn{2}{|c|}{0.601}&\multicolumn{2}{|c|}{0.223}&\multicolumn{2}{|c|}{0.652}
&\multicolumn{2}{|c|}{0.805}&\multicolumn{2}{|c|}{0.153}&\multicolumn{2}{|c|}{0.785}
&\multicolumn{2}{|c|}{0.706}&\multicolumn{2}{|c|}{0.158}&\multicolumn{2}{|c|}{0.714}
&\multicolumn{2}{|c|}{0.584}&\multicolumn{2}{|c|}{0.252}&\multicolumn{2}{|c|}{0.621}
\\
\multicolumn{4}{|c|}{\textbf{DRFI}~\cite{jiang2013salient}}
&\multicolumn{2}{|c|}{0.618}&\multicolumn{2}{|c|}{0.207}&\multicolumn{2}{|c|}{0.670}
&\multicolumn{2}{|c|}{0.807}&\multicolumn{2}{|c|}{0.148}&\multicolumn{2}{|c|}{0.797}
&\multicolumn{2}{|c|}{0.745}&\multicolumn{2}{|c|}{0.133}&\multicolumn{2}{|c|}{0.750}
&\multicolumn{2}{|c|}{0.634}&\multicolumn{2}{|c|}{0.224}&\multicolumn{2}{|c|}{0.624}
\\
\multicolumn{4}{|c|}{\textbf{DSR}~\cite{li2013saliency}}
&\multicolumn{2}{|c|}{0.558}&\multicolumn{2}{|c|}{0.215}&\multicolumn{2}{|c|}{0.594}
&\multicolumn{2}{|c|}{0.791}&\multicolumn{2}{|c|}{0.158}&\multicolumn{2}{|c|}{0.736}
&\multicolumn{2}{|c|}{0.712}&\multicolumn{2}{|c|}{0.141}&\multicolumn{2}{|c|}{0.715}
&\multicolumn{2}{|c|}{0.596}&\multicolumn{2}{|c|}{0.234}&\multicolumn{2}{|c|}{0.596}
\\
\hline
\end{tabular}
\vspace{-2mm}
\caption{Quantitative comparison with 15 methods on 4 complex structure image datasets. The best three results are shown in \textcolor[rgb]{1,0,0}{red},~\textcolor[rgb]{0,1,0}{green} and \textcolor[rgb]{0,0,1}{blue}, respectively. ``--'' means corresponding methods are trained on that dataset. Our method ranks first or second on these datasets.}
\label{table:fauc2}
\end{center}
\vspace{-4mm}
\end{table*}
%
{\flushleft\textbf{Quantitative Evaluation.}}
As illustrated in Tab.~\ref{table:fauc1}, Tab.~\ref{table:fauc2} and Fig.~\ref{fig:PR-curve}, our method outperforms other competing ones across all datasets in terms of near all evaluation metrics.
%
From these results, we have other notable observations: (1) deep learning based methods consistently outperform traditional methods with a large margin, which further proves the superiority of deep features for SOD.
(2) our method achieves higher S-measure than other methods, especially on complex structure datasets, \emph{e.g.}, the DUT-OMRON, SED and SOD datasets.
We attribute this result to our structural loss.
(3) without segmentation pre-training, our method only fine-tuned from the image classification model still achieves better results than the DCL and RFCN, especially on the HKU-IS and SED datasets.
%
(4) compared to the DHS and Amulet, our method is inferior on the DUTS-TE and PASCAL-S datasets.
However, our method ranks in the second place and is still very comparable.
\begin{figure*}
\centering
\begin{tabular}{@{}c@{}c@{}c@{}c@{}c@{}c@{}c@{}c@{}c@{}c@{}c@{}c}
\vspace{-0.5mm}
\hspace{-2mm}
\includegraphics[width=0.09\linewidth,height=1.35cm]{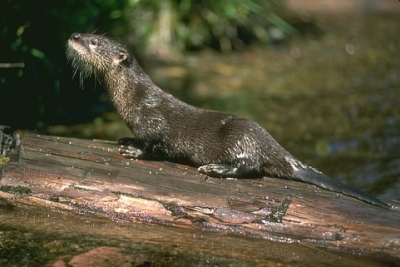}\hspace{0.1mm}\ &
\includegraphics[width=0.09\linewidth,height=1.35cm]{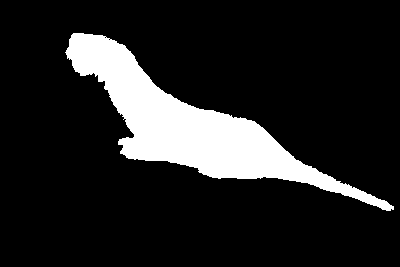}\hspace{0.1mm}\ &
\includegraphics[width=0.09\linewidth,height=1.35cm]{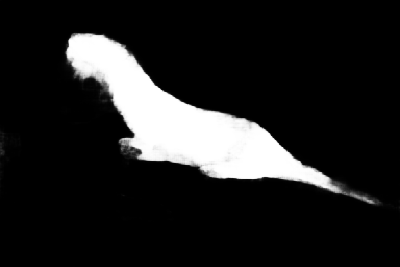}\hspace{0.1mm}\ &
\includegraphics[width=0.09\linewidth,height=1.35cm]{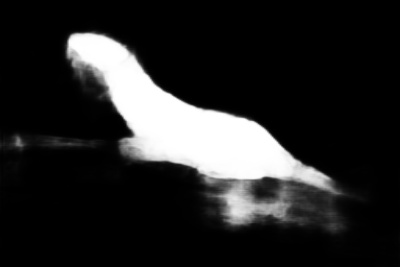}\hspace{0.1mm}\ &
\includegraphics[width=0.09\linewidth,height=1.35cm]{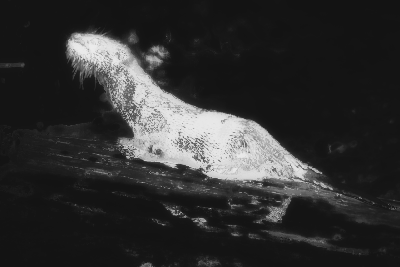}\hspace{0.1mm}\ &
\includegraphics[width=0.09\linewidth,height=1.35cm]{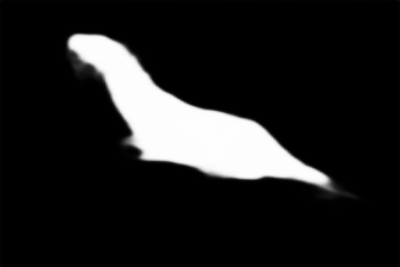}\hspace{0.1mm}\ &
\includegraphics[width=0.09\linewidth,height=1.35cm]{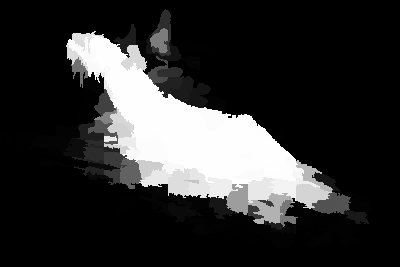}\hspace{0.1mm}\ &
\includegraphics[width=0.09\linewidth,height=1.35cm]{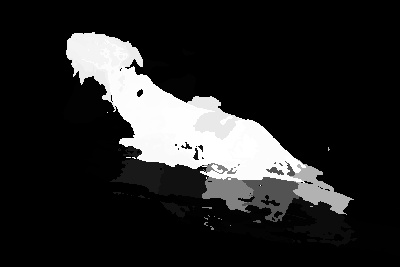}\hspace{0.1mm}\ &
\includegraphics[width=0.09\linewidth,height=1.35cm]{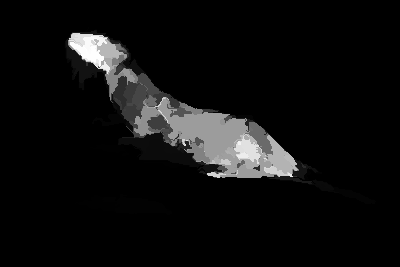}\hspace{0.1mm}\ &
\includegraphics[width=0.09\linewidth,height=1.35cm]{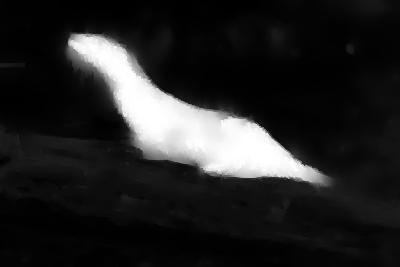}\hspace{0.1mm}\ &
\includegraphics[width=0.09\linewidth,height=1.35cm]{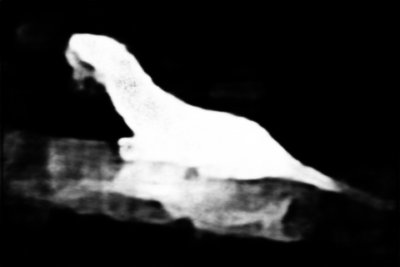}\hspace{0.1mm}\ \\
\vspace{-0.5mm}
\hspace{-2mm}
\includegraphics[width=0.09\linewidth,height=1.35cm]{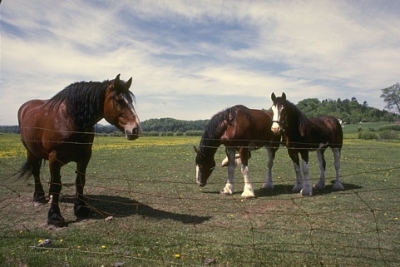}\hspace{0.1mm}\ &
\includegraphics[width=0.09\linewidth,height=1.35cm]{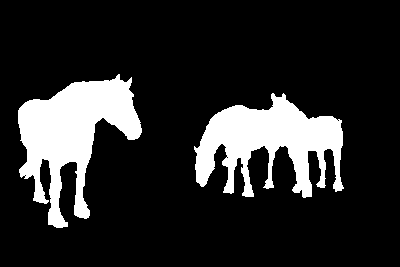}\hspace{0.1mm}\ &
\includegraphics[width=0.09\linewidth,height=1.35cm]{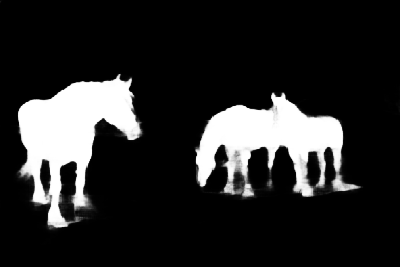}\hspace{0.1mm}\ &
\includegraphics[width=0.09\linewidth,height=1.35cm]{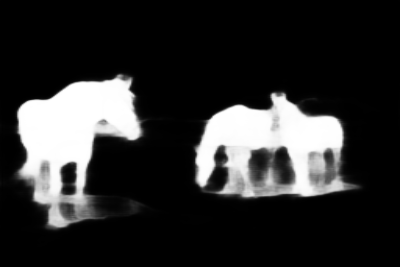}\hspace{0.1mm}\ &
\includegraphics[width=0.09\linewidth,height=1.35cm]{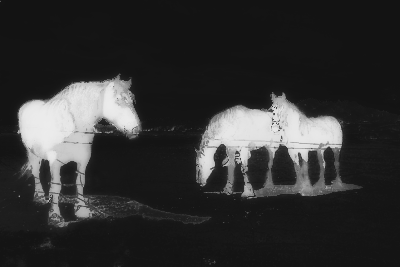}\hspace{0.1mm}\ &
\includegraphics[width=0.09\linewidth,height=1.35cm]{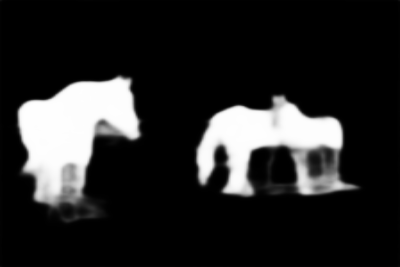}\hspace{0.1mm}\ &
\includegraphics[width=0.09\linewidth,height=1.35cm]{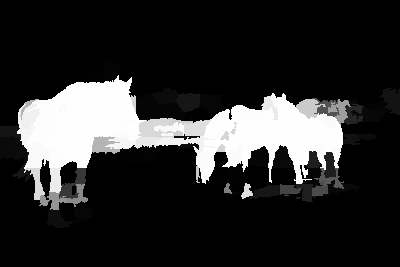}\hspace{0.1mm}\ &
\includegraphics[width=0.09\linewidth,height=1.35cm]{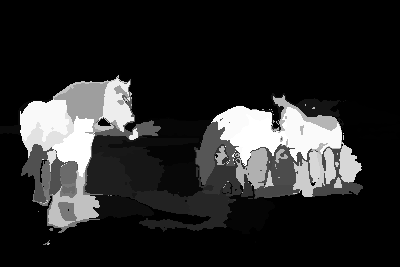}\hspace{0.1mm}\ &
\includegraphics[width=0.09\linewidth,height=1.35cm]{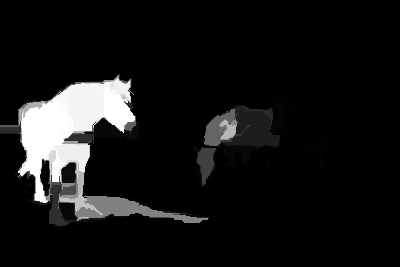}\hspace{0.1mm}\ &
\includegraphics[width=0.09\linewidth,height=1.35cm]{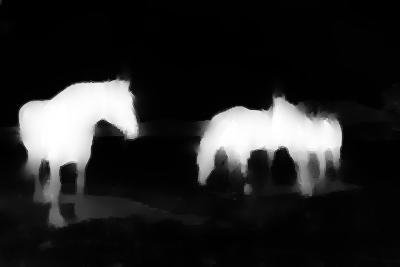}\hspace{0.1mm}\ &
\includegraphics[width=0.09\linewidth,height=1.35cm]{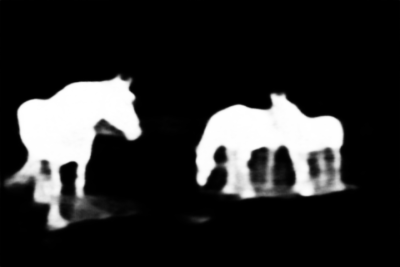}\hspace{0.1mm}\ \\
\vspace{-0.5mm}
\hspace{-2mm}
\includegraphics[width=0.09\linewidth,height=1.35cm]{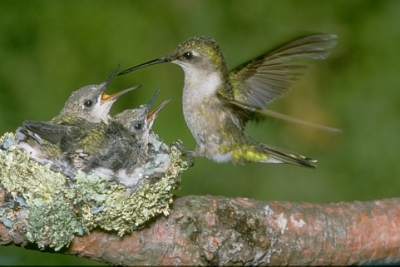}\hspace{0.1mm}\ &
\includegraphics[width=0.09\linewidth,height=1.35cm]{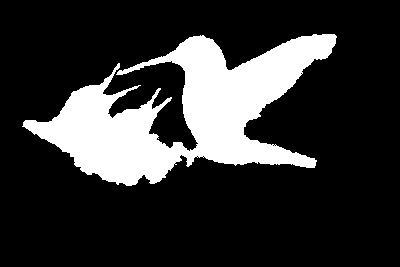}\hspace{0.1mm}\ &
\includegraphics[width=0.09\linewidth,height=1.35cm]{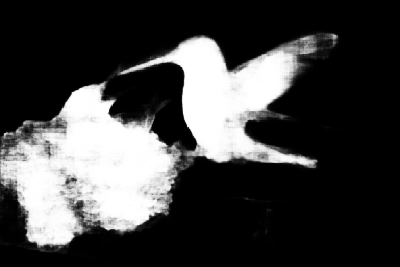}\hspace{0.1mm}\ &
\includegraphics[width=0.09\linewidth,height=1.35cm]{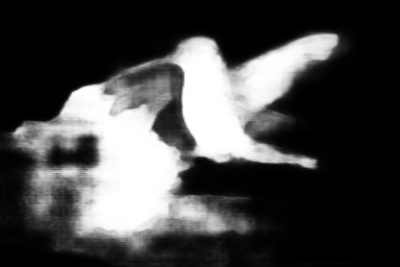}\hspace{0.1mm}\ &
\includegraphics[width=0.09\linewidth,height=1.35cm]{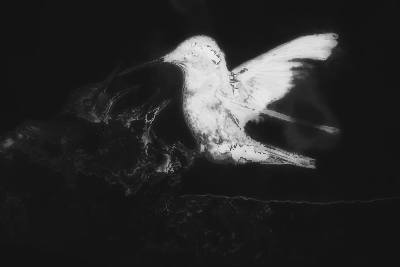}\hspace{0.1mm}\ &
\includegraphics[width=0.09\linewidth,height=1.35cm]{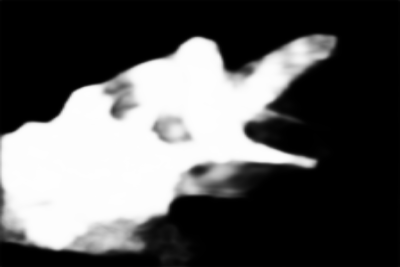}\hspace{0.1mm}\ &
\includegraphics[width=0.09\linewidth,height=1.35cm]{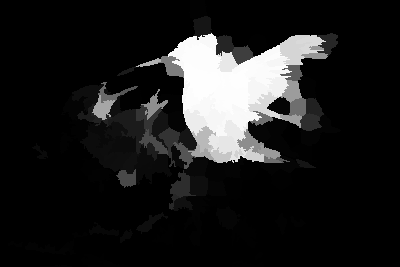}\hspace{0.1mm}\ &
\includegraphics[width=0.09\linewidth,height=1.35cm]{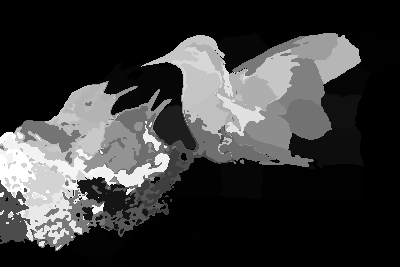}\hspace{0.1mm}\ &
\includegraphics[width=0.09\linewidth,height=1.35cm]{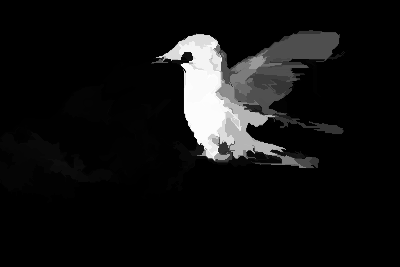}\hspace{0.1mm}\ &
\includegraphics[width=0.09\linewidth,height=1.35cm]{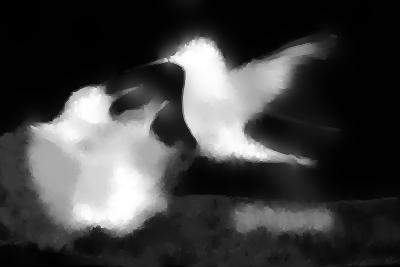}\hspace{0.1mm}\ &
\includegraphics[width=0.09\linewidth,height=1.35cm]{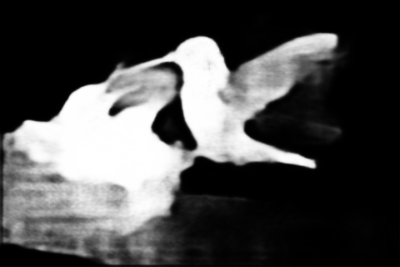}\hspace{0.1mm}\ \\
\vspace{-0.5mm}
\hspace{-2mm}
\includegraphics[width=0.09\linewidth,height=1.35cm]{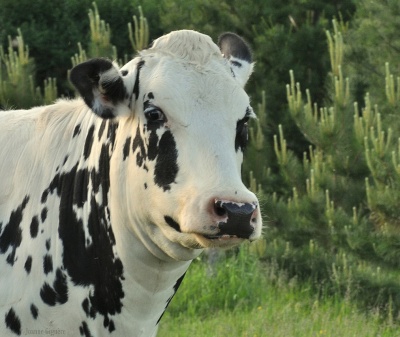}\hspace{0.1mm}\ &
\includegraphics[width=0.09\linewidth,height=1.35cm]{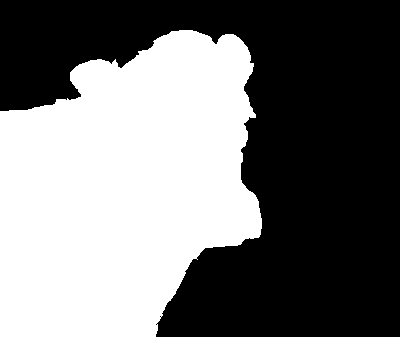}\hspace{0.1mm}\ &
\includegraphics[width=0.09\linewidth,height=1.35cm]{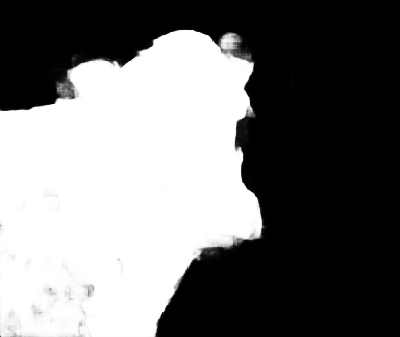}\hspace{0.1mm}\ &
\includegraphics[width=0.09\linewidth,height=1.35cm]{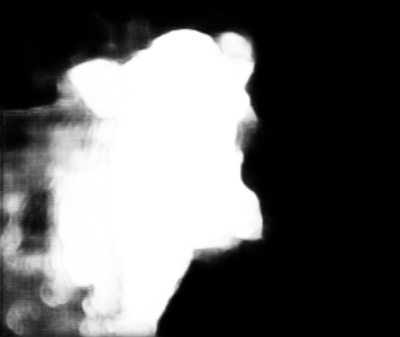}\hspace{0.1mm}\ &
\includegraphics[width=0.09\linewidth,height=1.35cm]{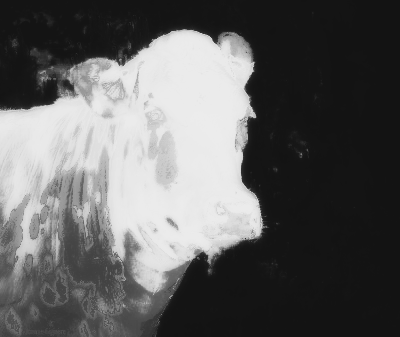}\hspace{0.1mm}\ &
\includegraphics[width=0.09\linewidth,height=1.35cm]{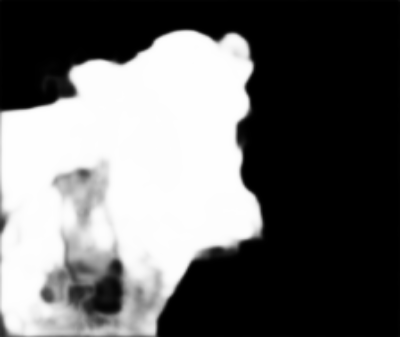}\hspace{0.1mm}\ &
\includegraphics[width=0.09\linewidth,height=1.35cm]{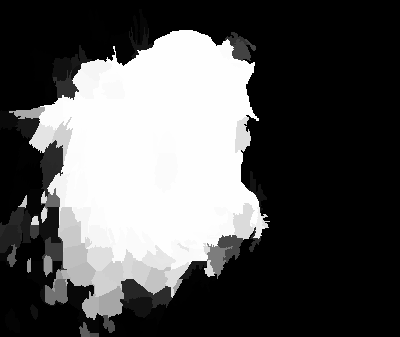}\hspace{0.1mm}\ &
\includegraphics[width=0.09\linewidth,height=1.35cm]{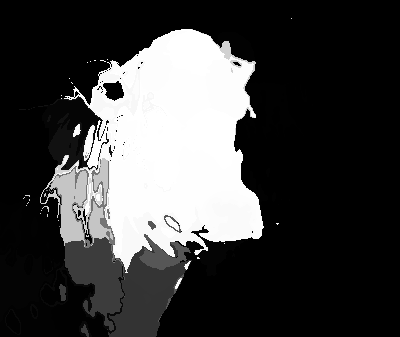}\hspace{0.1mm}\ &
\includegraphics[width=0.09\linewidth,height=1.35cm]{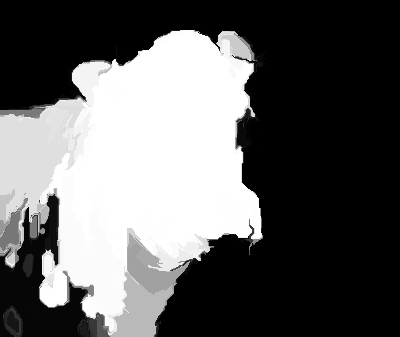}\hspace{0.1mm}\ &
\includegraphics[width=0.09\linewidth,height=1.35cm]{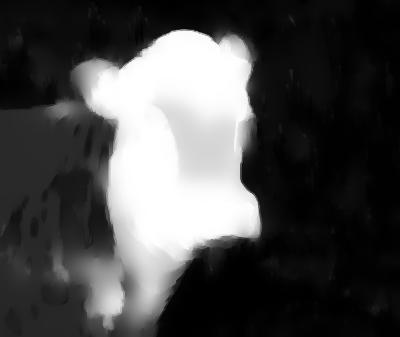}\hspace{0.1mm}\ &
\includegraphics[width=0.09\linewidth,height=1.35cm]{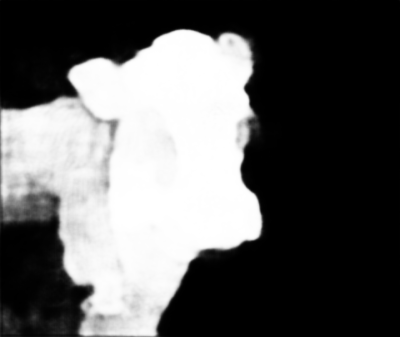}\hspace{0.1mm}\ \\
{\small (a)} & {\small(b)} & {\small(c)} & {\small(d)} & {\small(e)}& {\small(f)}& {\small(g)}
& {\small(h)}& {\small(i)}& {\small(j)}& {\small(k)}\ \\
\end{tabular}
\vspace{-2mm}
\caption{Comparison of typical saliency maps. (a) Input images; (b) Ground truth; (c) \textbf{Ours}; (d) \textbf{Amulet}; (e) \textbf{DCL}; (f) \textbf{DHS}; (g) \textbf{ELD}; (h) \textbf{MCDL}; (i) \textbf{MDF}; (j) \textbf{RFCN}; (k) \textbf{UCF}. Due to the limitation of space, we don't show the results of \textbf{DS}, \textbf{LEGS}, \textbf{BL}, \textbf{BSCA}, \textbf{DRFI} and \textbf{DSR}. We will release the saliency maps of all compared methods upon the acceptance.
\label{fig:map_comparison}}
\vspace{-4mm}
\end{figure*}
\begin{table*}
\begin{center}
\begin{tabular}{|c|c|c|c|c|c|c|}
\hline
Models &(a) \small{SFCN}-hf+$\mathcal{L}_{bce}$&(b) \small{SFCN}+$\mathcal{L}_{bce}$&(c) \small{SFCN}+$\mathcal{L}_{wbce}$&(d) \small{SFCN}+$\mathcal{L}_{wbce}$+$\mathcal{L}_{sc}$&(e) \small{SFCN}+$\mathcal{L}_{wbce}$+$\mathcal{L}_{s1}$&The overall \\
\hline
$F_\eta$       &0.824&0.848& 0.865& \textcolor[rgb]{0,1,0}{0.873}& \textcolor[rgb]{0,0,1}{0.867}&\textcolor[rgb]{1,0,0}{0.880}        \\
\hline
$MAE$          &0.102&0.083& 0.072& \textcolor[rgb]{0,0,1}{0.061}& \textcolor[rgb]{1,0,0}{0.049}&\textcolor[rgb]{0,1,0}{0.052}       \\
\hline
$S_\lambda$    &0.833&0.859& 0.864& \textcolor[rgb]{0,0,1}{0.880}& \textcolor[rgb]{0,1,0}{0.882}&\textcolor[rgb]{1,0,0}{0.897}       \\
\hline
\end{tabular}
\end{center}
\vspace{-4mm}
\caption{Results with different model settings on the ECSSD dataset. The best three results are shown in \textcolor[rgb]{1,0,0}{red},~\textcolor[rgb]{0,1,0}{green} and \textcolor[rgb]{0,0,1}{blue}, respectively.}
\label{table:aggregation}
\vspace{-6mm}
\end{table*}
{\flushleft\textbf{Qualitative Evaluation.}}
Fig.~\ref{fig:map_comparison} provides several visual examples in various challenging cases, where our method outperforms other compared methods.
For example, the images in the first two rows are of very low contrast, where most of the compared methods fail
to capture the salient objects, while our method successfully highlights them with sharper edges preserved.
The images in the 3-4 rows are challenging with complex structures or salient objects near the image boundary, and most of the compared methods can not predict the whole objects, while our method captures the whole salient regions with preserved structures.
\subsection{Ablation Analysis}
We also evaluate the main components in our model.
Tab.\ref{table:aggregation} shows the experimental results with different model settings.
All models are trained on the augmented MSRA10K dataset and share the same hyper-parameters described in subsection 4.2.
Due to the limitation of space, we only show the results on the ECSSD dataset.
Other datasets have the similar performance trend.
From the results, we can see that the SFCN only using the channel concatenation operator without hierarchical fusion (model (a)) has achieved comparable performance to most deep learning methods.
This confirms the effectiveness of reflection features.
With the hierarchical fusion, the resulting SFCN (model (b)) improves the performance by a large margin.
The main reason is that the fusion method introduces more contextual information from high layers to low layers, which helps to locate the salient objects.
In addition, it's no wonder that training with the $\mathcal{L}_{wbce}$ loss achieves better results than $\mathcal{L}_{bce}$.
With other two losses $\mathcal{L}_{sc}$ and $\mathcal{L}_{s1}$, the model achieves better performance in terms of MAE and S-measure.
These results demonstrate that individual components in our model complement each other.
When taking them together, the overall model, \emph{i.e.}, $SFCN+L_{wbce}+ L_{se} + L_{s1}$, achieves best results in all evaluation metrics.
\section{Conclusion}
In this work, we propose a novel end-to-end feature learning framework for SOD.
Our method uses a symmetrical FCN to learn complementary visual features under the guidance of lossless feature reflection.
For training, we also propose a new weighted structural loss that integrates the location, semantic and contextual information of salient objects to boost the detection performance.
Extensive experiments on seven large-scale saliency datasets demonstrate that the proposed method achieves significant improvement over the baseline and performs better than other state-of-the-art methods.
{\small {\flushleft\textbf{Acknowledgment}}.
This work is in part supported by the National Natural Science Foundation of China (NNSFC), No. 61472060, No. 61502070 and No. 61528101.
PP. Zhang and Wei Liu are currently visiting the University of Adelaide, supported by the China Scholarship Council (CSC) program.
This work is done during the visiting and supervised by Prof. Chunhua Shen.
}
\bibliographystyle{named}
\footnotesize
\bibliography{ijcai18}

\end{document}